\documentclass[10pt,twocolumn,letterpaper]{article}

\usepackage[pagenumbers]{cvpr} 

\definecolor{cvprblue}{rgb}{0.21,0.49,0.74}
\usepackage[pagebackref,breaklinks,colorlinks,allcolors=cvprblue]{hyperref}

\usepackage{cutwin}
\usepackage{enumitem}
\usepackage{amsmath}
\usepackage{booktabs}
\usepackage{multirow}
\usepackage{color}
\usepackage{amsfonts}
\usepackage{nicefrac}
\usepackage{microtype}
\usepackage{xcolor}
\usepackage{pifont}
\usepackage{graphicx} 
\usepackage{array} 
\usepackage{colortbl}
\usepackage{rotating}

\definecolor{LightCyan}{rgb}{0.88,1,1}

\usepackage{adjustbox}
\usepackage{makecell}
\usepackage{marvosym, ifsym} 

\definecolor{sgreen}{RGB}{30, 150, 30} 

\definecolor{amethyst}{rgb}{0.6, 0.4, 0.8}
\NewDocumentCommand{\Clawer}{ mO{} }{\textcolor{amethyst}{\textsuperscript{\textit{Clawer}}\textsf{{\small[{\it #1}]}}}}

\newcommand{\pipeline}{TextPecker}

\def\paperID{*****}
\def\confName{CVPR}
\def\confYear{2026}

\title{
  \raisebox{-3mm}{\includegraphics[width=0.065\textwidth]{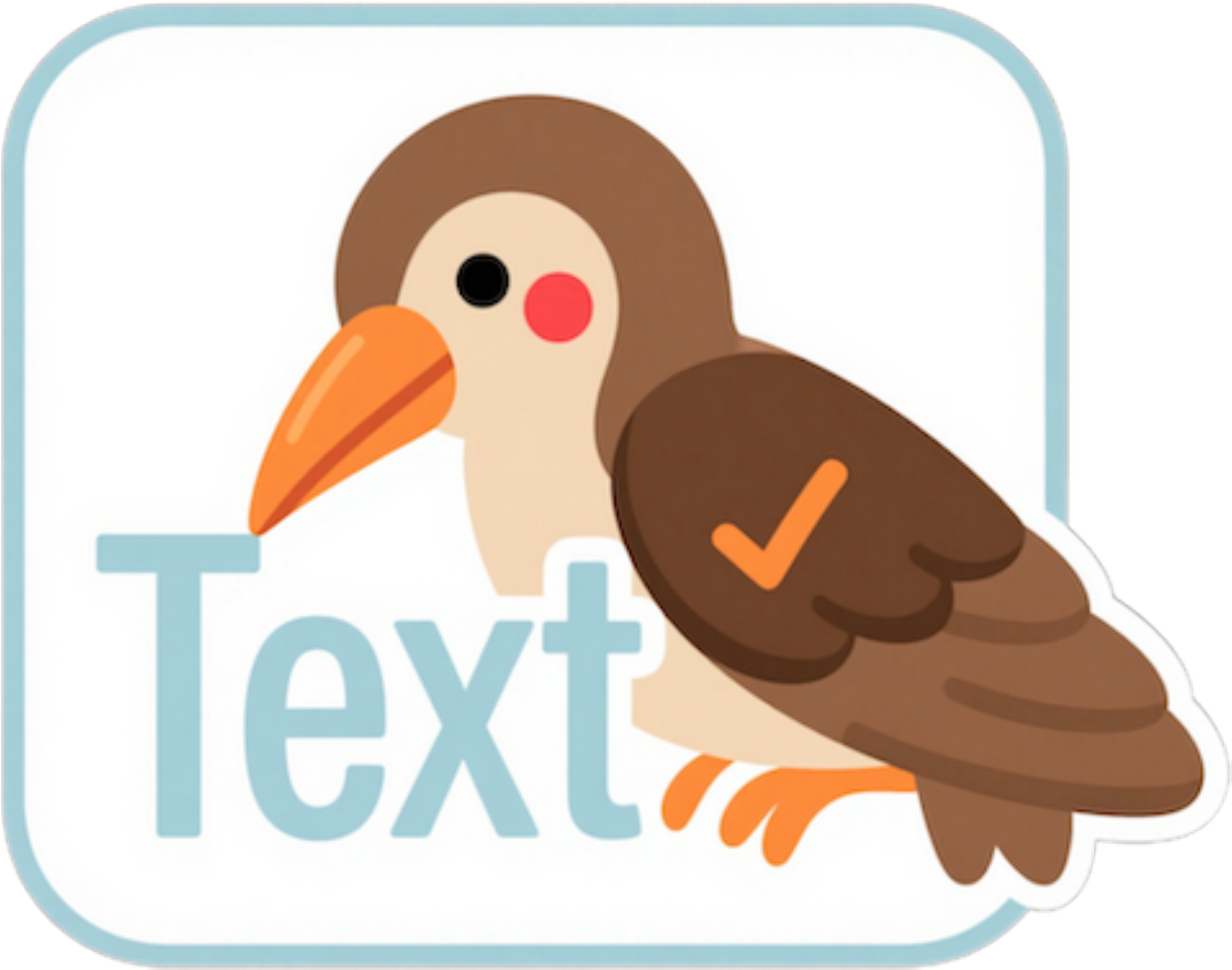}}TextPecker: Rewarding Structural Anomaly Quantification \\ \hspace{0.10em} for Enhancing Visual Text Rendering}

\author{
Hanshen Zhu$^{1,*}$,
Yuliang Liu$^{1}$,
Xuecheng Wu$^{2}$,
An-Lan Wang$^{2}$,  
Hao Feng$^{2}$, \\
Dingkang Yang$^{2}$,
Chao Feng$^{2}$,
Can Huang$^{2}$,
Jingqun Tang$^{2,\dagger}$, 
Xiang Bai\textsuperscript{1,\Letter} \\
$^1$Huazhong University of Science and Technology \quad 
$^2$ByteDance \\
{\tt\small \{zhs, ylliu, xbai\}@hust.edu.cn,} 
{\tt\small jingquntang@bytedance.com} \\
\vspace{0.1em}
\tt\small
{ \url{https://github.com/CIawevy/TextPecker}}
}

\begin{document}
\maketitle

\begin{abstract}
Visual Text Rendering (VTR) remains a critical challenge in text‑to‑image generation, where even advanced models frequently produce text with structural anomalies such as distortion, blurriness, and misalignment.
However, we find that leading MLLMs and specialist OCR models largely fail to perceive these structural anomalies, creating a critical bottleneck for both VTR evaluation and RL‑based optimization.   
As a result, even state‑of‑the‑art generators (\eg, Seedream4.0, Qwen‑Image) still struggle to render structurally faithful text.
To address this, we propose \textbf{\pipeline{}},
a plug-and-play structural anomaly perceptive RL strategy that mitigates noisy reward signals and works with any text-to-image generator. 
To enable this capability, we construct a recognition dataset with character‑level structural‑anomaly annotations and develop a stroke‑editing synthesis engine to expand structural‑error coverage. 
Experiments show that TextPecker consistently improves diverse text‑to‑image models; even on the well‑optimized Qwen‑Image, it significantly yields average gains of 4\% in structural fidelity and 8.7\% in semantic alignment for Chinese text rendering, establishing a new state-of-the-art in high-fidelity VTR.
Our work fills a gap in VTR optimization, providing a foundational step towards  reliable and structural faithful visual text generation.
\end{abstract}

\makeatletter
\renewcommand*{\@makefnmark}{}
\footnotetext{
$^*$Part of this work was done during Hanshen Zhu's internship at ByteDance. $^\dagger$Project leader. \textsuperscript{\Letter}Corresponding author. }

\section{Introduction}
\label{sec:intro}

\begin{figure}[tb]
  \centering
  \includegraphics[width=0.95\linewidth]{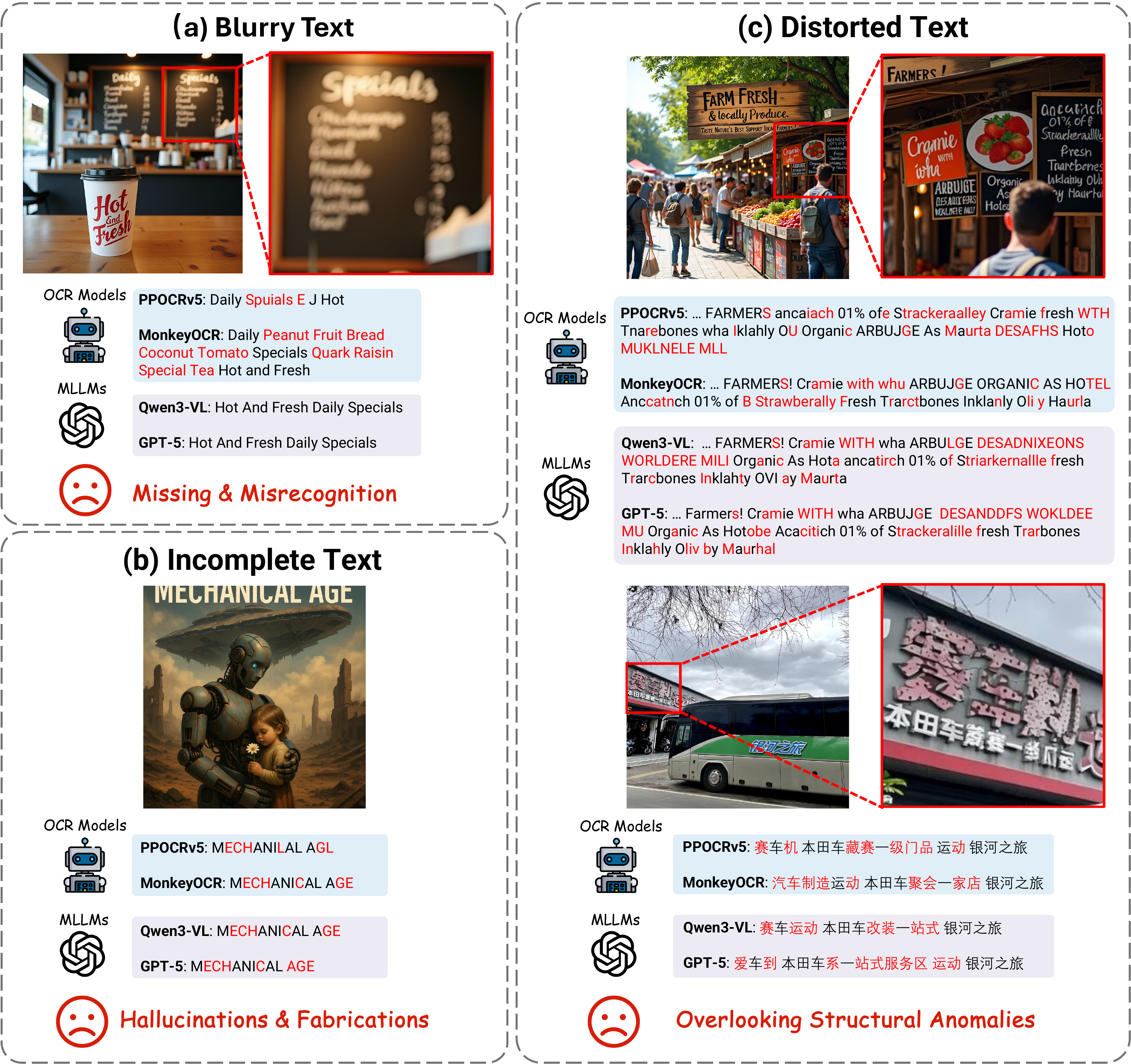}
  \vspace{-5pt}
    \caption{Existing OCR models and MLLMs struggle to perceive fine-grained structural anomalies in rendered text images, creating a key bottleneck for both VTR evaluation and RL-based optimization. Misrecognized characters are highlighted in \textcolor{red}{RED}.}
  \label{fig:challenges}
  \vspace{-10pt}
\end{figure}

Text-to-image generation has witnessed remarkable progress in producing photorealistic and detail-rich results \cite{SDXL,dalle3,imagen,Nano-Banana}. With these advancements, Visual Text Rendering (VTR), the task of generating legible and semantically consistent text within images, has emerged as a challenging and evolving frontier \cite{aestheticsIsCheap,QwenImage,Seeddream4.0}. However, the recent surge of specialized image generators (\eg, Flux-series \cite{flux}) and unified generative models (\eg, GPT-4o \cite{gpt4o}, BAGEL \cite{Bagel}) still struggle with VTR tasks, producing visual text with distortion, blurriness, misalignment, or missing characters \cite{aestheticsIsCheap}.
Prior works \cite{X-Omni_RL_and_LongText,QwenImage,Seedream3.0,blip3o-next} mitigate these issues with reinforcement learning (RL): the generated text is first recognized through OCR models \cite{PPOCRV5,got_ocr2.0} or Multimodal Large
Language Models (MLLMs) \cite{Qwen2.5-VL,qwen3-235b-a22b-ins,internvl3}, then rule-based scores (\eg, edit distance to the prompt) are computed and used as rewards. 
To evaluate text rendering performance, existing metrics \cite{X-Omni_RL_and_LongText,TextCrafter_cvtg2k,chang2025oneig} follow an analogous paradigm. This prevailing paradigm, however, rests on a flawed premise.

We identify a critical \textbf{bottleneck} shared by both VTR evaluation and reinforcement learning process: \textbf{a lack of fine-grained structural anomaly perception in rendered text}. As illustrated in Fig.~\ref{fig:challenges}, OCR models \cite{PPOCRV5,got_ocr2.0} and MLLMs \cite{Qwen2.5-VL,gpt5} are inherently ill-suited for this task.
Their failures manifest in two primary ways: (1) \textbf{Misinterpretation}: They over-rely on linguistic priors to ``correct'' or hallucinate semantic content from structurally flawed text, thereby ignoring subtle glyph-level defects (\eg, stroke deletions, misalignments, or spurious attachments). (2) \textbf{Invisibility}: They often fail to detect or simply dismiss low-confidence text regions, such as those with significant blurriness or distortion, treating them as non-existent.

Therefore, evaluators yield unreliable text‑accuracy estimates and fail to assess structural quality, and reward signals become misleading. As a direct result, even state-of-the-art generators (\eg, Qwen-Image \cite{QwenImage}, Seedream4.0 \cite{Seeddream4.0}) still struggle to render structurally faithful text. Our quantitative analysis in Tab.~\ref{tab:SAP_and_Rec} further substantiates this issue.

Building on these insights, we propose \textbf{TextPecker}, a plug-and-play structural anomaly perceptive RL strategy for visual text rendering. 
At its core, \pipeline{} replaces noisy OCR-based rewards with a perception-guided composite reward that jointly captures semantic alignment and structural fidelity.
Its structural term is sensitive to fine-grained glyph deformation and distortion, assigning reliable penalties to subtle defects that deceive structure-blind OCR and destabilize policy learning. This yields stable credit assignment and seamlessly integrates into any text-to-image generator without architectural changes.
To construct this reward, we address the scarcity of fine-grained structural annotations by building a hybrid dataset that couples two complementary sources:
(1) images with authentic generative artifacts from various text-to-image models, meticulously annotated at the character-level, and (2) synthetic data from our stroke-editing engine, crafted to expand error diversity while including normal characters for robust recognition.

Extensive experimental results demonstrate that our introduced TextPecker delivers consistent and significant improvements across diverse generators, including FLUX \cite{flux}, SD3.5 \cite{SD3.5}, and Qwen-Image \cite{QwenImage}. Remarkably, For FLUX, our method yields dramatic gains over its base version (\eg +38.3\% Sem. and +31.6\% Qua.) while also substantially outperforming the OCR-reward baseline. This advancement becomes even more pronounced on the highly-optimized Qwen-Image. In the challenging domain of Chinese text rendering, our approach achieves gains of 8.7\% in semantic alignment and 4\% in structural fidelity, establishing a new state-of-the-art in high-fidelity VTR.

\vspace{2pt}\noindent We summarize our {\bf contributions} as follows: \begin{itemize} 
\item We identify a critical \textbf{bottleneck} in VTR: the lack of fine-grained structural perception  in current OCR-based evaluators, which hinders effective VTR optimization.
\item We propose \textbf{TextPecker}, a plug-and-play structural anomaly perceptive RL strategy that seamlessly integrates into any text-to-image generators.
\item We construct a large-scale dataset with character-level structural anomaly annotations, addressing the data scarcity and enabling fine-grained structural perception for reward modeling.
\item Our method consistently improves leading generators and sets a new state-of-the-art in VTR, with notable gains even on the highly optimized Qwen-Image.
\end{itemize}

\section{Related Work}
\label{sec:related-work}
\subsection{Visual Text Rendering}
Text images are unique and crucial information media in modern digital society.
Mainstream text rendering methods broadly fall into two categories. The first focuses on specialized modules to incorporate additional constraints, such as glyph information \cite{anytext2,glyphcontrol,Glyphdraw2,brushyourtext} 
for text morphology control, or layout guidelines \cite{textdiffuser,dreamtext,postermaker,textctrl} for precise text placement.
The second focuses on improving text encoder designs. 
A prevailing explanation for the text rendering’s difficulty is that the text information often degrades or is inadequately preserved during encoding. 
To mitigate this, subsequent works \cite{udifftext,textdiffuser2,Glyph-byt5-v2} introduce special tokens for the rendered text or adopt tokenizer-free encoders like ByT5 \cite{byt5}.
With recent advances in generative models, specialized generators \cite{SD3.5,flux,Seeddream4.0,QwenImage} and unified genertative models \cite{Bagel,wu2025omnigen2,lin2025uniworld} already exhibit substantial text rendering capabilities without relying on ad-hoc designs such as glyph conditions.
Despite these improvements, they still struggle with text rendering tasks, producing visual text with distortion, blurriness, misalignment, or missing characters \cite{aestheticsIsCheap}.


\begin{figure*}[t!]
\centering
\includegraphics[width=\linewidth]{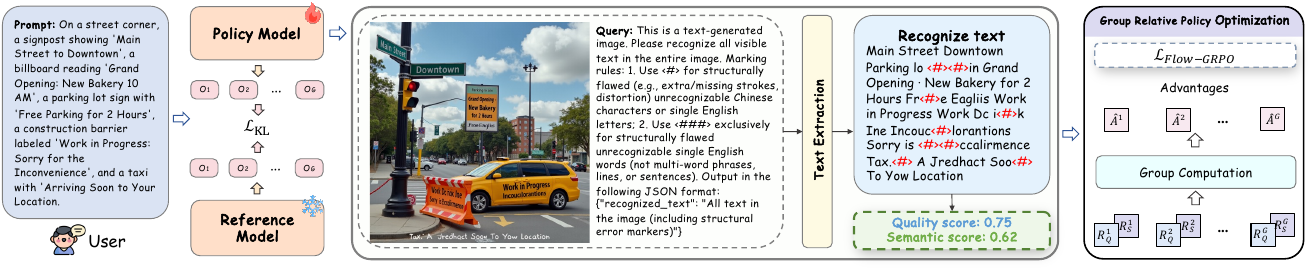}
\vspace{-15pt}
\caption{\textbf{Schematic illustration of the TextPecker framework}. Given a generative prompt, we first sample \( G \) candidate outputs \( \{o_i\}_{i=1}^G \) from the reference policy model \( \pi_{\theta_{\text{ref}}} \). Each \( o_i \) is sent to a structure-aware recognizer to extract fine-grained generated text, with markers indicating structurally anomalous text. We then compute the joint reward \( \mathcal{R}_i \), comprising a weighted sum of semantic alignment and structural quality scores (Sec.~\ref{sec:reward_modeling}). Each \( \mathcal{R}_i \) is normalized to a group relative advantage \( A_i \). Finally, we optimize the current policy model \( \pi_\theta \) by maximizing \( A_i \) while enforcing proximity to \( \pi_{\text{ref}} \) via KL divergence.}
\label{fig:method}
\vspace{-10pt}
\end{figure*}

\subsection{Evaluations for VTR}
The assessment of VTR has predominantly centered on textual accuracy~\cite{TextAtlas5M,chang2025oneig,TextCrafter_cvtg2k,X-Omni_RL_and_LongText}. A key challenge, particularly in non-glyph-conditioned generation, is the potential order mismatch between generated and target text. 
To address this, Lex-Art\cite{Lex_art_10k} proposed Pairwise Normalized Edit Distance (PNED), which combines Hungarian matching \cite{kuhn1955hungarian,semiets} with a penalty for unmatched words. TIIF‑Bench added global normalization to yield GNED\cite{TIIF-bench}, reducing sensitivity to text‑length imbalance.
Fang et al. \cite{fang2025flux-reason-6m} directly use Qwen2.5-VL \cite{Qwen2.5-VL} as end-to-end VTR evaluators, yet it suffers from hallucination and inaccuracies. 
Notably, recent works have also recognized the importance of structural quality of text. Font-Agent \cite{fontAgent} presents a stroke-aware font quality assessor, yet it is limited to single-character evaluations. He et al. \cite{he2025seeing} addresses OCR hallucinations in degraded documents, but is confined to document analysis domain.

Reward models are crucial for aligning generative models with human preferences during post-training optimization. Numerous efforts have largely improved reward models for general visual quality assessment \cite{HPSV2,unifiedReward,visualqualityR1}. Unlike subjective protocols such as aesthetic or quality, existing VTR reward modeling \cite{Seedream3.0,seedream2.0,X-Omni_RL_and_LongText,blip3o-next} has primarily focused on textual accuracy, which is typically assessed by \textit{de facto} evaluators such as standard OCR models \cite{PPOCRV5,got_ocr2.0} or MLLMs like Qwen2.5-VL \cite{Qwen2.5-VL}.

However, these methods for both evaluation and reward modeling are fundamentally constrained by their inability to perceive fine-grained structural anomalies, leading to unreliable accuracy estimates and misleading rewards.
In response, TextPecker's fine-grained, perception-guided composite reward moves beyond the noisy signals of OCR-based methods, enabling the joint quantification of semantic accuracy and structural fidelity.

\section{Methodology}

\subsection{Preliminaries}
\label{sec:prelim} 
Reinforcement learning (RL) is a common and effective paradigm for improving text rendering in text-to-image models \cite{flux,QwenImage,Seeddream4.0}, with widely adopted variants such as DPO \cite{DPO} and GRPO \cite{deepseek}. We focus on GRPO, a critic-free on-policy method that stabilizes policy optimization via group-wise relative advantages and intra-group reward normalization. However, due to the deterministic nature of flow-matching \cite{rectifiedflow,flowmatching} models, they are not intrinsically designed for reinforcement learning.
Flow-GRPO \cite{flowgrpo} extends GRPO to the rectified-flow setting by injecting stochasticity into the integration process. Specifically, it converts the deterministic dynamics into a stochastic differential equation:
\begin{equation}
dx_t = \left( v_t + \frac{\sigma_t^2}{2t}\left( x_t + (1-t)v_t \right) \right) dt + \sigma_t \, dw_t,
\label{eq:flow_grpo_update}
\end{equation}
where \(v_t = v_\theta(x_t, t, c)\) is the network-predicted velocity, \(dw_t\) denotes Brownian motion, and \(\sigma_t = a\sqrt{\tfrac{t}{1-t}}\) controls the magnitude of stochasticity.

For VTR Optimization, prior works \cite{flowgrpo,Seedream3.0,X-Omni_RL_and_LongText,blip3o-next} predominantly leverage a string-level accuracy reward: $ S = 1 - N_e/N_t $, where $N_t$ denotes the length of target string and $N_e$ is the edit distance between the target string and the OCR-extracted content. As discussed earlier (cf. Fig.~\ref{fig:challenges}), existing OCR Models \cite{PPOCRV5,got_ocr2.0} and MLLMs \cite{Qwen2.5-VL} prioritize semantic recovery over glyph integrity, hallucinating corrections for structurally flawed text and omitting low-confidence distorted regions. Such behaviors depress the edit distance $N_e$ and inflate the reward score $S$, resulting in biased rewards that hinder effective optimization.

\subsection{TextPecker}
\label{sec:textpecker}
As shown in Fig.~\ref{fig:method}, we introduce \textbf{TextPecker}, a plug-and-play RL strategy for enhancing VTR with fine-grained structural perception. Unlike conventional methods that rely on noisy OCR signals and overlook structural flaws, TextPecker redefines reward modeling by jointly optimizing semantic alignment and structural fidelity. Integrating this perception-guided composite reward into the RL loop yields consistent gains across diverse generators.

\subsubsection{Structure-aware Reward Functions}
\label{sec:reward_modeling}

To address the structural blindness and overconfident scoring of prior OCR-based methods, our reward formulation is built upon a \textbf{structure-aware assessment module}. As illustrated in Fig.~\ref{fig:method}, this module identifies fine-grained structural anomalies in the generated text (\eg, missing or spurious strokes) and flags them with special markers. The detailed construction of this module is presented in Sec.~\ref{sec:data_cons}. Assuming we have such a module, we formulate our composite reward as follows.

\paragraph{Structural Quality Score ($\mathcal{S}_Q$).}
An intuitive way to quantify structural quality is to measure the proportion of ``bad'' characters. We define the structural quality score, $\mathcal{S}_Q$, based on the ratio of structurally anomalous characters to the total number of characters. However, for powerful generators, structural errors are often rare but visually jarring when they do occur. To amplify the penalty for such infrequent yet critical failures, we introduce a scaling factor $\omega > 1$. The final score is thus formulated as:
\begin{equation}
\mathcal{S}_Q=\mathrm{clip}\big(1-\omega\,\tfrac{N_a}{N_P},\,0,\,1\big),
\end{equation}
where $N_P$ is the total number of characters in the generated text $\mathcal{P}$, and $N_a$ is the number of characters flagged as anomalous by our assessor. Here, $\mathrm{clip}$ constrains the value to the range $[0, 1]$.

\paragraph{Semantic Alignment Score ($\mathcal{S}_E$).}
Unlike prior OCR-based rewards that treat text as a simple long string, we argue that word-level matching is crucial for accurately assessing text that may not be rendered in the same order as the prompt. Inspired by ~\cite{Lex_art_10k,TIIF-bench}, we also find it necessary to penalize any unmatched words, which typically include extraneous or repeated texts in the generated output, as well as missing textual content from the target prompt. Therefore, we formulate our semantic alignment score as:
\begin{equation}
\mathcal{S}_{E} = 1 - \frac{\sum_{(t_i, p_j) \in \mathcal{M}} \mathrm{NED}(t_i, p_j) + \text{Penalty}(\mathcal{T}, \mathcal{P}, \mathcal{M})}{\max(|\mathcal{T}|, |\mathcal{P}|)}.
\end{equation}
Here, $\mathcal{T}$ and $\mathcal{P}$ are the sets of words in the target and generated text, respectively. $\mathcal{M}$ represents the optimal word pairing between $\mathcal{T}$ and $\mathcal{P}$, found via the Hungarian algorithm to achieve the minimum alignment cost based on Normalized Edit Distance (NED). The $\text{Penalty}(\cdot)$ term is the count of any unmatched words. This ensures that both superfluous generated words and missing target words contribute to the overall error. The final score $\mathcal{S}_E$ is also clipped to the range of $[0, 1]$, where a higher value indicates better semantic alignment.

\paragraph{Composite Reward ($\mathcal{R}$).}
Finally, we formulate the overall TextPecker reward $\mathcal{R}$ as a weighted sum of the two scores, allowing for a joint optimization of both aspects:
\begin{equation}
\mathcal{R}=w_E\,\mathcal{S}_E+w_Q\,\mathcal{S}_Q,\qquad w_E+w_Q=1.
\end{equation}

\begin{figure}[t!]
\centering
\includegraphics[width=\linewidth]{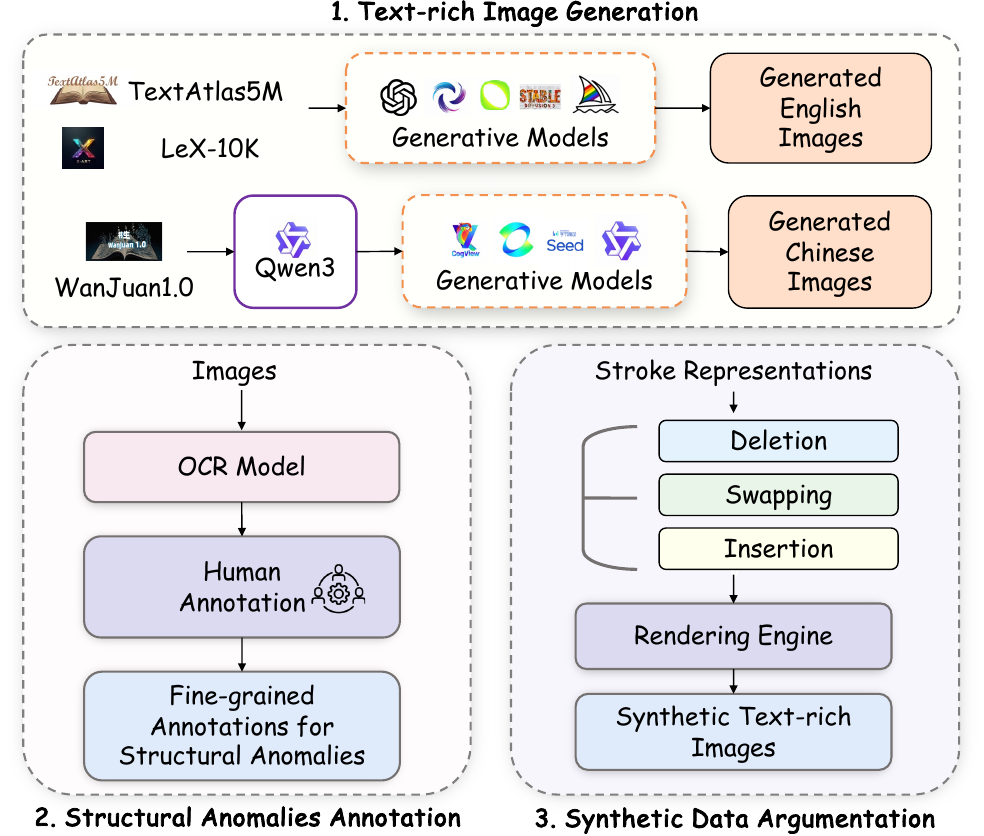}
\vspace{-15pt}
\caption{The illustration of proposed data construction pipeline.}
\label{fig:data_pipe}
\vspace{-15pt}
\end{figure}


\subsubsection{Structural Perceptive Data Construction}
\label{sec:data_cons}

Our structure-aware reward relies on a robust assessor for fine-grained structural anomalies, but the requisite labeled data is critically scarce. To construct a large-scale, high-quality dataset, we proceed in \textbf{three steps} (Fig.~\ref{fig:data_pipe}):

\vspace{1mm}\noindent\textbf{Step 1: Text-rich Image Generation.}
The initial step involves constructing a large-scale dataset of text-rich visual images, covering diverse structural error types.
Specifically, for English text generation, we draw prompts from TextAtlas5M \cite{TextAtlas5M} and Lex-10k \cite{Lex_art_10k}, and leverage multiple English-capable generative models (Anytext, Stable Diffusion v1-5 \cite{SD}, Stable Diffusion 3.5 \cite{SD3.5}, Flux \cite{flux}, Seedream3.0 \cite{Seedream3.0}, Qwen-Image \cite{QwenImage}) to generate such images. For Chinese text generation, we first sample a comprehensive text corpus from WanJuan1.0 \cite{wanjuan}, ensuring coverage of modern Chinese common characters
We then use Qwen3-235B-A22B \cite{qwen3-235b-a22b-ins} to generate descriptions of various font styles, which are integrated with the corpus to form final prompts for models including Cogview4 \cite{cogview}, Kolors \cite{Kolors}, Seedream3.0 \cite{Seedream3.0}, and Qwen-Image \cite{QwenImage}.

\vspace{1mm}\noindent\textbf{Step 2: Structural Anomaly Annotation.}
Generated text-rich images exhibit diverse structural anomalies. We define such anomalies as any structural distortion impairing semantic recognition, caused by blurring, warping, missing strokes, or redundant artifacts. To streamline annotation, we first leverage OCR models~\cite{PPOCRV5} to obtain preliminary recognition results. Annotators then identify and rectify fine-grained character-level structural flaws with \textbf{a special marker} (as illustrated in Fig.~\ref{fig:method}). For words with severe structural adhesion that prevents accurate character counting, we use a distinct placeholder, yielding a dataset with refined fine-grained labels for structural anomalies.

\vspace{1mm}\noindent\textbf{Step 3: Synthetic Data Augmentation.} 
While Step 2’s annotations capture common structural anomalies, models trained solely on them exhibit two key limitations: poor generalization to unseen anomalies and degraded recognition of Chinese characters (Tab.~\ref{tab:ablation_rec}). This stems from the intrinsic complexity of Chinese: unlike the linear morphology of English, Chinese characters have a 2D spatial composition and a vast inventory of over 8,000, causing a combinatorial explosion of structural anomalies beyond what exhaustive annotation can cover. To overcome this, we introduce a synthesis-based augmentation that programmatically generates diverse erroneous and canonical Chinese characters.

We start by representing Chinese characters as compositions of fundamental strokes, modeled as ordered sequences using stroke order data from public resources.
To streamline manipulation, we uniformly sample points along each stroke to enable manipulation. With stroke sequences and their point representations, we define three stroke-level structural edit operators. We apply them sequentially and compose them to produce diverse structural anomalies. (1)  \emph{Stroke Deletion:} removes a controlled subset of strokes. (2)  \emph{Stroke Swapping:} exchanges the locations of disjoint stroke pairs by aligning centroids. (3)  \emph{Stroke Insertion:} adds strokes sampled from other characters.
These operators generate structurally anomalous characters, we then build a rendering engine on top of SynthTIGER \cite{synthtiger} and use it to place structurally anomalous and canonical text onto diverse backgrounds and layouts, producing text-rich images. 
We merge the annotated and synthetic data to form the final training and test splits, with dataset statistics and visualized distributions shown in Tab.~\ref{tab:dataset_summary}. 






\begin{table}[tb]
\scriptsize
\setlength{\tabcolsep}{2.4mm}
\caption{Statistics of our constructed text-rich image recognition dataset with structural-anomaly labels at box and image levels. Proportions are computed over all instances.}
\label{tab:dataset_summary}
\renewcommand{\arraystretch}{1.1}
\centering
\begin{tabular}{c|c|c|c}
\toprule 
Data Type      & Level   & Samples  & Proportion \\
\midrule

\multirow{2}{*}{Manual Annotations}          
& Box & 559.6K    & 39.32\%     \\
& Image & 131.1K        & 9.21\%    \\
\midrule 

\multirow{2}{*}{Synthetic Anomaly Text}            
& Box & 452.5K    & 31.80\%     \\
& Image & 100.0K        & 7.03\%    \\
\midrule 

\multirow{2}{*}{Synthetic Normal Text} 
& Box & 150.0K    & 10.54\%     \\
& Image & 30.0K        & 2.10\%    \\
\midrule

\multirow{1}{*}{\textbf{Total}} 
& \textbf{--}   &  1.4M    & 100\%   \\
\bottomrule

\end{tabular}
\vspace{-10pt}
\end{table}


\section{Experiments}
\label{sec:exps}

\begin{table*}[tb]
\scriptsize
\setlength{\tabcolsep}{2.5mm}  
\caption{Results of Text Structural Anomaly Perception (TSAP) and Canonical Text Recognition (CTR): measuring model's recognition ability under generated text images and TSAP-demanding prompts. Box-level results of non-supporting models are marked as "-".}
\label{tab:SAP_and_Rec}
\renewcommand\arraystretch{1.0}
\centering
\begin{tabular}{l|ccc|ccc|cc|cc} 
\toprule
\multirow{3}{*}{Methods} & \multicolumn{6}{c|}{\textbf{TSAP}} & \multicolumn{4}{c}{\textbf{CTR}} \\  
\cmidrule(lr){2-7} \cmidrule(lr){8-11}
& \multicolumn{3}{c|}{Image-level} & \multicolumn{3}{c|}{Box-level} & \multicolumn{2}{c|}{Image-level} & \multicolumn{2}{c}{Box-level} \\  
\cmidrule(lr){2-4} \cmidrule(lr){5-7} \cmidrule(lr){8-9} \cmidrule(lr){10-11}
& P & R & F1 & P & R & F1 & R & NED & R & NED  \\  
\midrule
\multicolumn{11}{l}{\textbf{\emph{English recognition}}} \\
\midrule
PP-OCRv5 \cite{PPOCRV5} & 0.000 & 0.000 & 0.000 & - & - & - & 0.720 & 0.137 & - & -  \\
GOT-OCR-2.0 \cite{got_ocr2.0} & 0.000 & 0.000 & 0.000 & - & - & - & 0.610 & 0.186 & - & -  \\
MonkeyOCR \cite{monkeyocr} & 0.000 & 0.000 & 0.000 & - & - & - & 0.578 & 0.209 & - & -  \\
Gemini-2.5-pro \cite{gemini2.5pro} & 0.179 & 0.076 & 0.107 & 0.342 & 0.179 & 0.235 & 0.415 & 0.557 & 0.300 & 0.571  \\
Doubao-Seed-1.6 \cite{doubaoseed16} & 0.157 & 0.167 & 0.162 & 0.333 & 0.180 & 0.234 & 0.714 & 0.169 & 0.376 & 0.473  \\
Doubao-Seed-1.6-think \cite{doubaoseed16thinking} & 0.259 & 0.183 & 0.214 & 0.280 & 0.095 & 0.141 & 0.736 & 0.119 & 0.418 & 0.414  \\
GPT-5 \cite{gpt5} & 0.196 & 0.150 & 0.170 & 0.419 & 0.193 & 0.265 & 0.556 & 0.359 & 0.398 & 0.450  \\
Qwen3-VL-8B \cite{qwen3-235b-a22b-ins} & 0.286 & 0.017 & 0.032 & 0.500 & 0.018 & 0.034 & 0.807 & 0.078 & 0.761 & 0.112  \\
InternVL3-8B \cite{internvl3} & 0.206 & 0.165 & 0.183 & 0.218 & 0.443 & 0.293 & 0.759 & 0.102 & 0.551 & 0.302  \\
\midrule 
\rowcolor{gray!10}
TextPecker (InternVL3-8B) \cite{internvl3} & \textbf{0.795} & \underline{0.960} & \textbf{0.870} & \textbf{0.784} & \textbf{0.964} & \textbf{0.865} & \textbf{0.944} & \textbf{0.035} & \textbf{0.953} & \textbf{0.030}  \\
\rowcolor{gray!10}
TextPecker (Qwen3-VL-8B) \cite{qwen3-235b-a22b-ins} & \underline{0.777} & \textbf{0.969} & \underline{0.862} & \underline{0.714} & \underline{0.938} & \underline{0.811} & \underline{0.918} & \underline{0.046} & \underline{0.941} & \underline{0.038}   \\
\midrule
\multicolumn{11}{l}{\textbf{\emph{Chinese recognition}}} \\
\midrule
PP-OCRv5 \cite{PPOCRV5} & 0.300 & 0.013 & 0.024 & - & - & - & 0.921 & 0.067 & - & -  \\
GOT-OCR-2.0 \cite{got_ocr2.0} & 0.500 & 0.004 & 0.008 & - & - & -  & 0.853 & 0.136 & - & -   \\
MonkeyOCR \cite{monkeyocr} & \textbf{1.000}  & 0.013 & 0.025 & - & - & - & 0.917  & 0.076 & - & -  \\
Gemini-2.5-pro \cite{gemini2.5pro} & 0.079 & 0.048 & 0.059 & 0.333 & 0.099 & 0.152 & 0.574 & 0.422 & 0.526 & 0.473  \\
Doubao-Seed-1.6 \cite{doubaoseed16} & 0.306 & 0.182 & 0.228 & 0.310 & 0.115 & 0.168 & 0.918 & 0.079 & 0.677 & 0.322  \\
Doubao-Seed-1.6-think \cite{doubaoseed16thinking} & 0.216 & 0.084 & 0.121 & 0.333 & 0.086 & 0.137 & 0.915 & 0.080 & 0.758 & 0.242  \\
GPT-5 \cite{gpt5} & 0.233 & 0.220 & 0.226 & 0.282 & 0.165 & 0.209 & 0.758 & 0.239 & 0.729 & 0.270  \\
Qwen3-VL-8B\cite{qwen3-235b-a22b-ins} & 0.400 & 0.008 & 0.017 & 0.000 & 0.000 & 0.000 & 0.943 & 0.054 & 0.931 & 0.068  \\
InternVL3-8B  \cite{internvl3} & 0.190 & 0.128 & 0.153 & 0.129 & 0.370 & 0.191 & 0.927 & 0.069 & 0.729 & 0.270  \\
\midrule 
\rowcolor{gray!10}
TextPecker (InternVL3-8B) \cite{internvl3} & \underline{0.889} & \underline{0.968} & \textbf{0.927} & \textbf{0.912} & \textbf{0.988} & \textbf{0.949} & \underline{0.962} & \underline{0.037} & \textbf{0.991} & \textbf{0.009}  \\
\rowcolor{gray!10}
TextPecker (Qwen3-VL-8B) \cite{qwen3-235b-a22b-ins} & 0.874 & \textbf{0.981} & \underline{0.925} & \underline{0.900} & \underline{0.964} & \underline{0.931} & \textbf{0.972} & \textbf{0.027} & \underline{0.989} & \underline{0.010} \\
\bottomrule
\end{tabular}
\vspace{-2pt}
\end{table*}

\subsection{Experimental settings}
Additional implementation details of our method, Baselines, dataset, and metrics are provided in the Appendix.

\vspace{1mm} \noindent \textbf{Implementation Details.}  
We adopt Qwen3-VL-8B \cite{qwen3-235b-a22b-ins} and InternVL3-8B \cite{internvl3} as the base architecture for \pipeline{}, leveraging their strong general recognition performance, robust cross-modal alignment, and native support for boundary box input. Fully supervised fine-tuning uses a batch size of 2, gradient accumulation steps of 32, learning rate of 5e-6, warm-up ratio of 0.05, and runs for 2 epochs.  
For VTR optimization, we employ Flow-GRPO \cite{flowgrpo} and validate on three popular models: SD3.5-M \cite{SD3.5}, Flux.1[dev] \cite{flux}, and Qwen-Image \cite{QwenImage}. We set $\omega=5$ and $w_E = w_Q=0.5$  and adopt the Qwen3-VL-based variant for reward function. Following Flow-GRPO \cite{flowgrpo}, other hyperparameters vary across models but are consistent within each; full details are provided in the Appendix.
Both the recognizer training and VTR optimization experiments are conducted on 32 NVIDIA H20 GPUs.

\vspace{1mm} \noindent \textbf{Metrics and Datasets.}  
We design two tasks, Text Structural Anomaly Perception (\emph{TSAP}) and Canonical Text Recognition (\emph{CTR}), to evaluate models’ fine-grained recognition ability under generated text images. TSAP measures whether the predicted anomalous character count \( N_a \) lies within the interval \([\delta \times N_a', N_a'/\delta]\), where \(\delta\) is a tolerance hyperparameter set to 0.7, with such interval-matched cases used to compute Precision, Recall, and F1-scores. For CTR, we report two metrics: Recall and NED for fine-grained assesment.
Evaluations are conducted on the test set proposed in \cref{sec:data_cons}.

For Text Rendering tasks, we first evaluate models on three established benchmarks: OneIG-Bench \cite{chang2025oneig}, CVTG-2K \cite{TextCrafter_cvtg2k}, and LongText-Bench \cite{X-Omni_RL_and_LongText}. We identified that their reliance on structurally-unaware OCR Models \cite{PPOCRV5,Qwen2.5-VL} can yield unreliable metrics, a critical limitation shared by a broader range of prominent benchmarks \cite{TextAtlas5M,TIIF-bench,Lex_art_10k}.
To address this, we first re-evaluate these benchmarks using \pipeline{}, reporting both semantic alignment scores and structural quality scores (defined in Sec.~\ref{sec:reward_modeling} with $\omega=1$ for evaluation).  
To streamline this re-evaluation with TextPecker and leverage the strengths of existing benchmarks, we integrate English and Chinese prompts curated from these datasets \cite{chang2025oneig,X-Omni_RL_and_LongText,TextCrafter_cvtg2k,TextAtlas5M,Lex_art_10k,QwenImage}, referred to as \emph{GenTextEval} for brevity. Given the scarcity of Chinese-rendering benchmarks \cite{chang2025oneig,X-Omni_RL_and_LongText,QwenImage}, we supplement this integrated set with Chinese prompts constructed in Sec.~\ref{sec:data_cons}, resulting in a total of 314 English prompts and 417 Chinese prompts. More details are provided in the Appendix.

\begin{table*}[tb]
\scriptsize
\setlength{\tabcolsep}{2.5mm}
\caption{Quantitative comparisons of different RL-optimized generative models on OneIG-Bench \cite{chang2025oneig}, LongText-Bench \cite{X-Omni_RL_and_LongText}, CVTG-2k \cite{TextCrafter_cvtg2k}, and GenTextEval-Bench. \textbf{Avg.}: Average text score from original benchmarks; \textbf{Qua.}: structural Quality score. \textbf{Sem.}: Semantic alignment. Score measurement and reward computation are both conducted by TextPecker (Qwen3-VL).}
\vspace{-6pt}
\label{tab:vtr_benchmarks}
\renewcommand\arraystretch{1.0}
\centering
\begin{tabular}{l|l|ccc|ccc|ccc|cc}
\toprule
\multirow{2}{*}{Models}  
& \multirow{2}{*}{Rewards}  
& \multicolumn{3}{c|}{OneIG} 
& \multicolumn{3}{c|}{LongText} 
& \multicolumn{3}{c}{CVTG-2K} 
& \multicolumn{2}{c}{GenTextEval} \\
\cmidrule(lr){3-5}  \cmidrule(lr){6-8}  \cmidrule(lr){9-11} \cmidrule(lr){12-13}
 & &   Avg. & 
Qua. & Sem.
&  Avg. &  Qua. &  Sem.
&  Avg. &  Qua. &  Sem.
 &  Qua. &  Sem. \\
\midrule
\multicolumn{11}{l}{\textbf{\emph{English Rendering}}} \\
\midrule
\multirow{3}{*}{SD3.5-M \cite{SD3.5}} & - & 0.441 &  0.848 & 0.513 & 0.296 & 0.860 & 0.416 & 0.368 &  0.869 & 0.491  & 0.671  & 0.265 \\
& OCR & 0.572 &  0.941 & 0.627 & 0.295 & 0.944 & 0.498  & 0.513 & 0.943  & \textbf{0.671}  & 0.940 & 0.462  \\
& TextPecker & \textbf{0.581} & \textbf{0.957} & \textbf{0.636} & \textbf{0.344} & \textbf{0.957} & \textbf{ 0.503} & \textbf{0.596} & \textbf{ 0.944} & 0.593  &  \textbf{0.959}&\textbf{0.506} \\
\midrule
\multirow{3}{*}{Flux.1[dev] \cite{flux}} & - & 0.567 &  0.875 &  0.585 & 0.613 &  0.929  & 0.591  & 0.491 & 0.908 & 0.523 & 0.672 & 0.336\\
& OCR & 0.754 &  0.969 & 0.708 & 0.736 & 0.972  & 0.707 & \textbf{0.778} & 0.951 & 0.632 &0.976 & 0.602   \\
& TextPecker & \textbf{0.845} & \textbf{ 0.979 } & \textbf{0.734} & \textbf{0.811} & \textbf{0.986} &0.672   & 0.777 & \textbf{0.961} & \textbf{0.704} & \textbf{0.988} & \textbf{0.719}\\
\midrule
\multirow{4}{*}{Qwen-Image \cite{QwenImage}} & - & 0.871 & 0.955 & 0.814 & 0.935 & 0.970 & 0.844 & 0.834 & 0.964  & 0.817  & 0.964 & 0.729 \\
& OCR & 0.986 & 0.983& 0.894 & \textbf{0.949} & 0.986  & 0.912 & 0.893 & 0.978 & 0.908 &  0.989 & 0.827 \\
& TextPecker & \textbf{0.990} & \textbf{0.988} & \textbf{0.910} & 0.945 & \textbf{0.990}& \textbf{0.918 } & \textbf{0.899}  & \textbf{0.987} & \textbf{0.932} & \textbf{0.992} & \textbf{0.837}\\
\midrule
\multicolumn{11}{l}{\textbf{\emph{Chinese Rendering}}} \\
\midrule
\multirow{4}{*}{Qwen-Image \cite{QwenImage}} & - & 0.954 & 0.894 & 0.732  & 0.920 & 0.924 & 0.834  & - & - &-   & 0.933 &0.810 \\
& OCR & 0.984 &  0.945 & 0.856 & 0.967 &  0.956 & 0.886  &- &- &-  &  0.953 &0.874\\
& TextPecker & \textbf{0.988} & \textbf{0.956} & \textbf{0.875} & \textbf{0.974} & \textbf{0.969} &\textbf{0.908} & - & - & - & \textbf{0.973}& \textbf{0.897}  \\
\bottomrule
\end{tabular}
\vspace{-10pt}
\end{table*}

\vspace{1mm} \noindent \textbf{Baselines.} \
We compare TextPecker against both specialist OCR models and general MLLMs. The former includes PPOCRv5 \cite{PPOCRV5}, GOT-OCR-2.0 \cite{got_ocr2.0}, MonkeyOCR \cite{monkeyocr}. For the latter, we benchmark against leading proprietary models such as GPT-5 \cite{gpt5}, Gemini-2.5-Pro \cite{gemini2.5pro}, and Doubao-Seed
-1.6 \cite{doubaoseed16,doubaoseed16thinking}, as well as strong open-source alternatives like Qwen3-VL \cite{qwen3-235b-a22b-ins} and InternVL3\cite{internvl3}.

\subsection{Main Results}
Additional results on evaluator generalization, RL baselines, multi-reward and ablations are presented in the appendix.

\begin{figure*}[htb]
  \centering
  \includegraphics[width=0.95\linewidth]{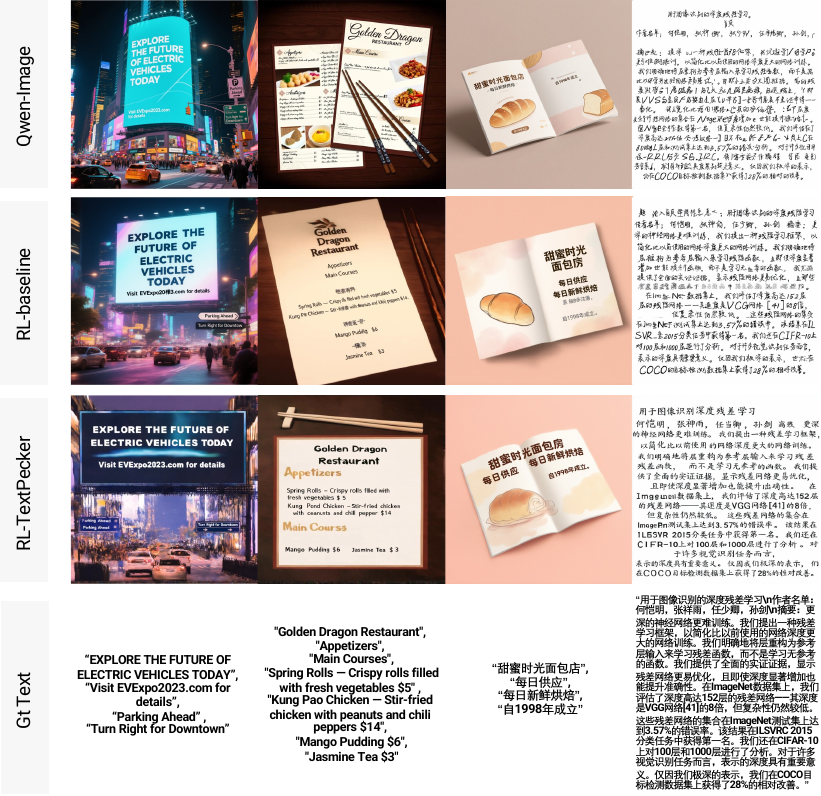}
  \vspace{-10pt}
   \caption{Qualitative Comparisons of Text Rendering for Qwen-Image and RL-Optimized Variants. Readers are highly recommended to refer to the appendix for extensive comparisons across generative models and RL baselines.}
  \label{fig:RL_EXP_QWEN}
  \vspace{-13pt}
\end{figure*}

\subsubsection{Quantitative Results}
\vspace{1mm} \noindent \textbf{TSAP and CTR.}
As illustrated in Tab.~\ref{tab:SAP_and_Rec}, our quantitative analysis reveals significant limitations in existing models and highlights the superiority of TextPecker.

First, we observe a near-total failure of both leading MLLMs and specialist OCR models on the Text Structural Anomaly Perception (TSAP) task. While a few MLLMs like Doubao-Seed-1.6 \cite{doubaoseed16}, GPT-5 \cite{gpt5}, and InternVL3 \cite{internvl3} show a nascent ability for structural anomaly perception, their performance remains rudimentary. This failure stems from a core mismatch in task nature: (1) these models are trained for generalized text recognition, which prioritizes robust semantic extraction over structural authenticity; for instance, when facing text partially visible due to occlusion, they are encouraged to predict the complete semantics corresponding to their expected full structure rather than verify structural fidelity (Fig.~\ref{fig:challenges}). Moreover, generated text frequently contains diverse structural anomalies that are rarely encountered in standard text recognition scenarios.

Besides, we identify a widespread deficiency in box-level text recognition. While traditional OCR models fundamentally lack the ability to process region-specific inputs, modern MLLMs also struggle with structural-anomaly detection. As shown in Tab.~\ref{tab:SAP_and_Rec}, their box-level recognition recall is substantially lower than their image-level counterparts. This limitation severely hinders more fine-grained assessments, which are crucial for evaluating demanding tasks such as controllable text editing and translation of text in local areas.
In stark contrast, TextPecker's models excel across all dimensions. They not only achieve high F1 and recall on the TSAP task but also improve recognition of generated text (CTR) compared to their respective baselines, underscoring the value of our box-level structure-aware dataset. Specifically, the Qwen3-VL-based variant attains state-of-the-art Chinese recognition performance, while the InternVL3-based variant exhibits the strongest overall capabilities at the box level.

\vspace{1mm} \noindent \textbf{RL for VTR.}
Tab.~\ref{tab:vtr_benchmarks} quantifies the effectiveness of our RL-based optimization across diverse scenarios. Overall, integrating TextPecker's structure-aware reward yields consistent improvements across four benchmarks, three base models, and both English and Chinese rendering tasks. For Flux.1[dev] \cite{flux} on English, gains are pronounced: +38.3\% Sem. and +31.6\% Qua. over the base model, and +11.7\% Sem. over the OCR-reward baseline on GenTextEval.
While gains are consistent overall, we also observe a few seemingly contradictory results when evaluated by different metrics. For example, SD3.5-M \cite{SD3.5} on CVTG-2K reports +8.3\% average word accuracy across five regions, yet -7.8\% Sem. compared to the OCR-reward baseline. This divergence indicates that structure-unaware, OCR-based metrics can overestimate performance on structurally flawed text, underscoring the necessity of a structure-aware assessor for faithful evaluation—despite these localized divergences, the overall benchmark averages still show marked improvement.
Notably, even for the well-optimized Qwen-Image \cite{QwenImage}, TextPecker-based RL delivers significant improvements, especially on Chinese rendering: +14.3\% on OneIG, +7.4\% on LongText, and +8.7\% on GenTextEval over prior SOTA; compared to the OCR-reward baseline, Qua. increases by +2.0\% and Sem. by +2.3\%. Taken together, these results establish TextPecker as a generalizable, plug-and-play reward that advances structurally faithful and accurate text rendering.

\subsubsection{Qualitative Results}
We present qualitative comparisons for Qwen‑Image and its RL‑optimized variants in Fig.~\ref{fig:RL_EXP_QWEN}. The vanilla Qwen‑Image, despite achieving prior SOTA on several text rendering benchmarks, often produces off‑target strings and exhibits blurred, distorted, or misaligned text, particularly in small and dense text regions. Optimizing with an OCR‑based reward helps reducing off‑target content and semantic alginment is enhanced, yet structural defects persist. In contrast, TextPecker‑based RL achieves superior structural fidelity and semantic consistency (\eg the English menu and Chinese paper cases). These observations align with the quantitative gains in Tab.~\ref{tab:vtr_benchmarks} and the component analysis in Tab.~\ref{tab:rl_ablation}, and we observe similar trends across other backbones (see Appendix), underscoring TextPecker’s effectiveness as a reward for structurally faithful and accurate text rendering.

In stark contrast, further optimization with TextPecker achieves a superior level of both structural fidelity and semantic consistency. This is particularly evident in challenging cases where OCR-based rewards falter. For instance, in the ``paper rendering'' case, our method successfully renders clean, aligned paragraphs where the OCR-rewarded model still produces distorted and wavy text lines. Similarly, in the ``English menu'' example, TextPecker accurately generates crisp, legible items that the baseline struggles to form correctly. 

\begin{table}[tb]
\tiny
\setlength{\tabcolsep}{0.9mm}
\caption{Ablation study on effectiveness of the constructed data: Image-level recognition results across different baseline models.}
\label{tab:ablation_rec}
\renewcommand\arraystretch{1.1}
\centering
\begin{tabular}{l|cc|ccc|cc|ccc|cc}   
\toprule
\multirow{3}{*}{Models} & \multicolumn{2}{c|}{Settings} & \multicolumn{5}{c|}{English} & \multicolumn{5}{c}{Chinese} \\
\cmidrule(lr){2-3} \cmidrule(lr){4-8} \cmidrule(lr){9-13}
& \multirow{2}{*}{Anno.} & \multirow{2}{*}{Syn.} & \multicolumn{3}{c|}{TSAP} & \multicolumn{2}{c|}{CTR} & \multicolumn{3}{c|}{TSAP} & \multicolumn{2}{c}{CTR} \\
\cmidrule(lr){4-6} \cmidrule(lr){7-8} \cmidrule(lr){9-11} \cmidrule(lr){12-13}
& & & P & R & F1 & R & NED & P & R & F1 & R & NED  \\
\midrule
\multirow{3}{*}{InternVL3 \cite{internvl3}} 
&  &  & 0.206 & 0.165 & 0.183 & 0.759 & 0.102 & 0.190 & 0.128 & 0.153 & \underline{0.927} & \underline{0.069} \\
& \checkmark &  & \textbf{0.795} & \textbf{0.970} & \textbf{0.874} & \underline{0.938} & \underline{0.042} & \underline{0.583} & \underline{0.966} & \underline{0.727} & 0.849 & 0.148 \\
& \checkmark & \checkmark & \underline{0.795} & \underline{0.960} & \underline{0.870} & \textbf{0.944} & \textbf{0.035} & \textbf{0.889} & \textbf{0.968} & \textbf{0.927} & \textbf{0.962} & \textbf{0.037} \\
\midrule
\multirow{3}{*}{Qwen3-VL \cite{qwen3-235b-a22b-ins}} 
&  &  & 0.182 & 0.018 & 0.032 & 0.810 & 0.076 & 0.333 & 0.008 & 0.017 & \underline{0.947} & \underline{0.049} \\
& \checkmark &  & \underline{0.706} & \underline{0.944} & \underline{0.808} & \underline{0.899} & \underline{0.068} & \underline{0.643} & \underline{0.942} & \underline{0.764} & 0.839 & 0.117 \\
& \checkmark & \checkmark & \textbf{0.777} & \textbf{0.969} & \textbf{0.862} & \textbf{0.918} & \textbf{0.046} & \textbf{0.874} & \textbf{0.981} & \textbf{0.925} & \textbf{0.972} & \textbf{0.027} \\
\bottomrule
\end{tabular}
\vspace{-12pt}
\end{table}

\subsection{Ablation Studies}
We conduct ablation studies to validate the effectiveness of our proposed dataset and to dissect the contribution of each component within the \pipeline{} reward mechanism.

\vspace{1mm} \noindent \textbf{Effectiveness of Data Composition.}
As shown in Tab.~\ref{tab:ablation_rec}, training on our annotated data alone yields substantial improvements on the TSAP task, with F1 scores for both models significantly outperforming the baseline. This data also enhances English text recognition. However, we observe a noticeable degradation in Chinese recognition performance, which we attribute to the increased complexity of structural anomalies in Chinese characters. The addition of our synthesized data effectively resolves this issue. By training on a combined dataset, both models demonstrate a dramatic boost in Chinese performance for both TSAP and CTR, achieving precise recognition and high accuracy in detecting structural anomalies. For English, the impact is model-dependent: while Qwen3-VL shows consistent enhancement, InternVL3 exhibits a slight decline in TSAP performance, yet gains notable improvements in CTR.

\vspace{1mm} \noindent \textbf{Analysis of Reward Components.}
We deconstruct the reward function step-by-step to isolate the impact of each component. As shown in Tab.~\ref{tab:rl_ablation}, combining a conventional, structure-unaware OCR model with Pairwise Matching (PM) yields a 4.2\% gain in the semantic alignment score, yet the structural quality score remains stagnant—indicating that GNED sharpens semantic feedback but fails to improve structural fidelity when the recognizer lacks structural perception. Replacing the OCR model with TextPecker delivers gains across both dimensions (Sem. +5.8\%, Qua. +0.8\%), as TextPecker’s reward is \textbf{inherently structure-aware}: characters identified with special markers directly modulate the semantic score. Finally, incorporating the structural quality term as an auxiliary reward brings further improvements and achieves the best overall performance, confirming the synergy of the full TextPecker reward design.

\begin{table}[tb]
\scriptsize
\setlength{\tabcolsep}{0.9mm}  
\caption{Ablation study on the effectiveness of reward design:  PM for Pair-wise Matching, and SQ for Structural Quality reward.}
\label{tab:rl_ablation}
\renewcommand\arraystretch{1.1}
\centering
\begin{tabular}{l|c|ccc|cc}   
\toprule
\multirow{2}{*}{Generative Model} & \multirow{2}{*}{OCR Model} & \multicolumn{3}{c|}{Settings} & \multicolumn{2}{c}{GenTextEval-EN} \\
\cmidrule(lr){3-5} \cmidrule(lr){6-7}  
& & NED & PM & SQ & Qua. & Sem. \\
\midrule
\multirow{5}{*}{Flux.1[dev] \cite{flux}} 
& - &   &  &  & 0.672 & 0.336 \\
& PP-OCRv5 \cite{PPOCRV5} &\checkmark   &  &  & 0.976 & 0.602 \\
& PP-OCRv5 \cite{PPOCRV5} & \checkmark  & \checkmark &  & 0.976 & 0.644 \\
& TextPecker & \checkmark  & \checkmark &  & 0.984 & 0.702 \\
& TextPecker & \checkmark   & \checkmark & \checkmark & \textbf{0.988} & \textbf{0.719} \\
\bottomrule
\end{tabular}
\vspace{-12pt}
\end{table}

\section{Conclusion}
\label{sec:conclusion}
We pinpoint and address the core bottleneck in VTR evaluation and RL-based optimization: leading MLLMs and specialist OCR models largely fail to perceive fine‑grained structural anomalies. We present TextPecker, a plug-and-play, structural-anomaly–aware framework that couples a standardized evaluator with an RL reward, providing complementary reward signals for semantic alignment and structural quality.
Empirically, TextPecker delivers consistent gains across leading text-to-image generators, including notable improvements on the well-optimized Qwen-Image. Under structure-aware rewards, generation behavior shifts toward fewer off-target strings, reduced blur and distortion, and improved alignment. This work provides a foundational step towards structurally faithful visual text rendering and supplies the community with essential tools for rigorous evaluation and post-training enhancement.




{
\small
\bibliographystyle{ieeenat_fullname}
\bibliography{main_bk}
}

\clearpage 

\twocolumn[
  \begin{center}
    \begin{minipage}{0.9\textwidth}
      \centering
      \Large\bfseries TextPecker: Rewarding Structural Anomaly Quantification \\ for Enhancing Visual Text Rendering \\
      \large (Supplementary Materials)
    \end{minipage}
  \end{center}
  \vspace{1.5em} 
]

\appendix 


\section{Additional Results}\label{supp:sec:more_results}  
\begin{figure*}[t!]
  \centering
  \includegraphics[width=\linewidth]{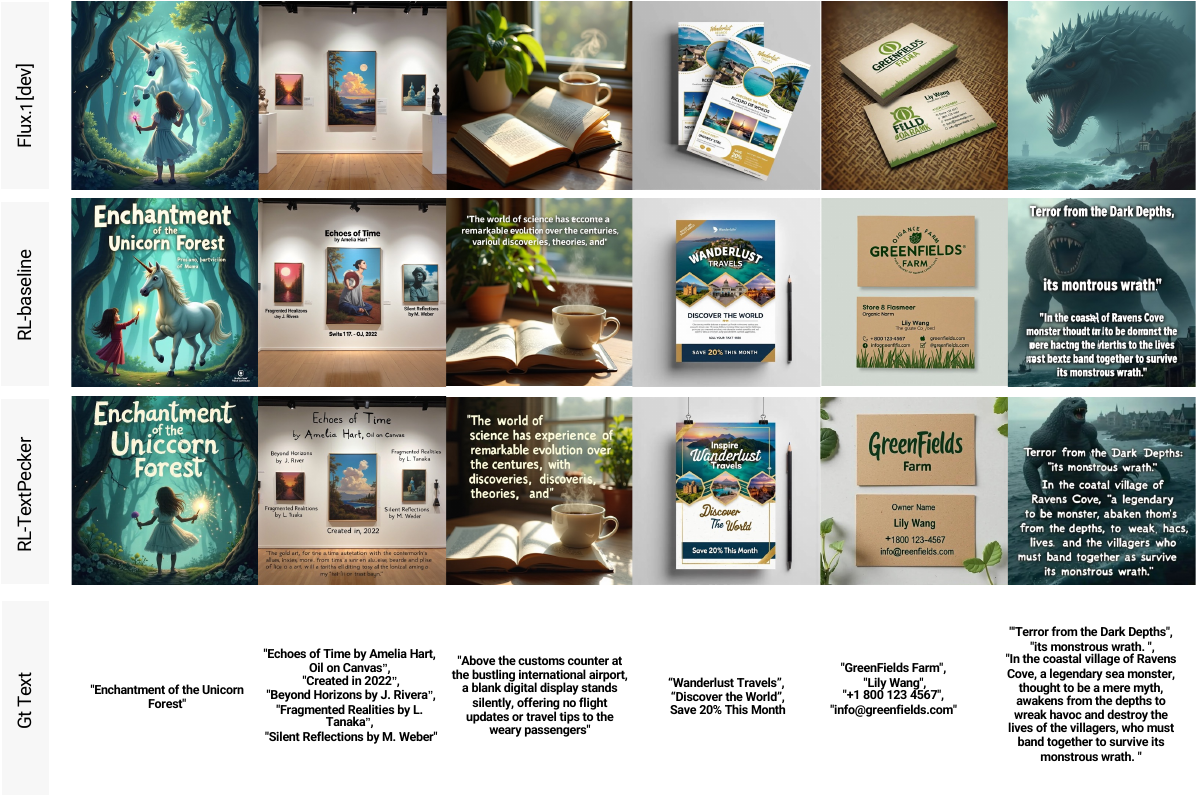}
  \vspace{-5pt}
   \caption{Qualitative Comparisons of Text Rendering for Flux.1[dev] \cite{flux} and RL-Optimized Variants.}
  \label{fig:RL_EXP_Flux}
  \vspace{-10pt}
\end{figure*}

\begin{figure*}[t!]
\centering
\includegraphics[width=\linewidth]{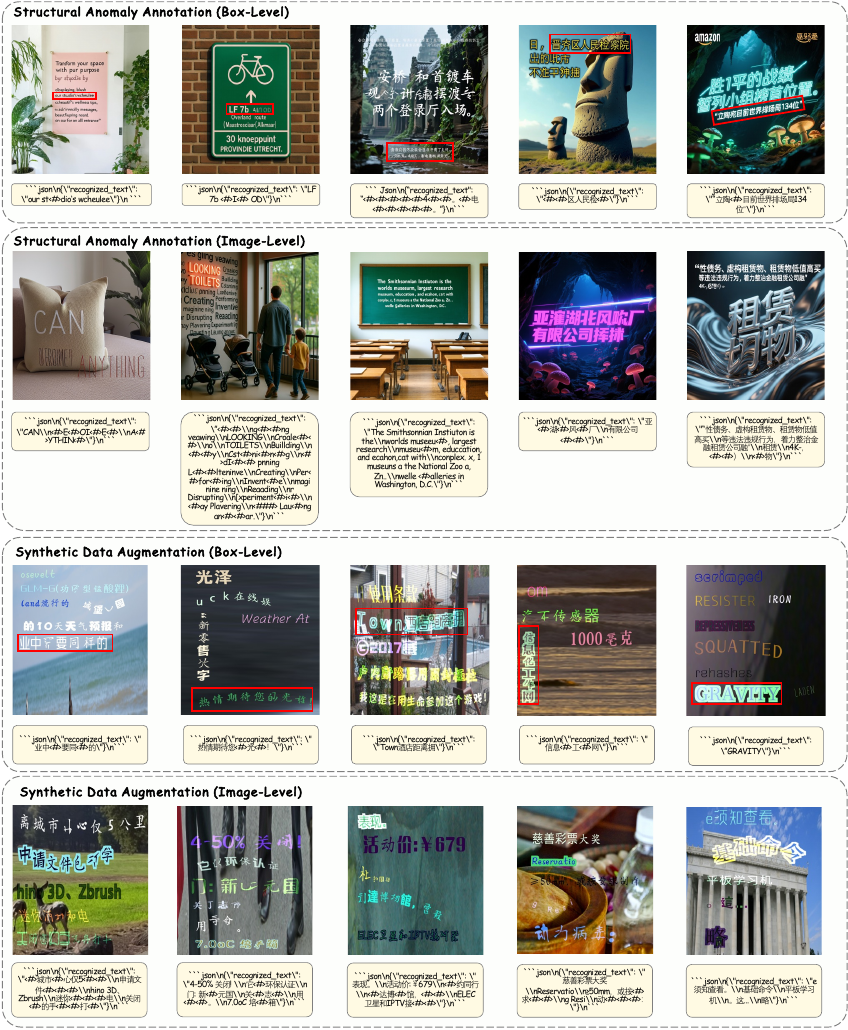}
\vspace{-15pt}
\caption{This figure shows manually annotated structural anomalies (box-level and image-level) from various generative models, alongside synthetic structural anomalies generated by our rendering engine for data augmentation.}
\label{supp:fig:data_vis}
\vspace{-10pt}
\end{figure*}

\begin{figure*}[t!]
\centering
\includegraphics[width=0.95\linewidth]{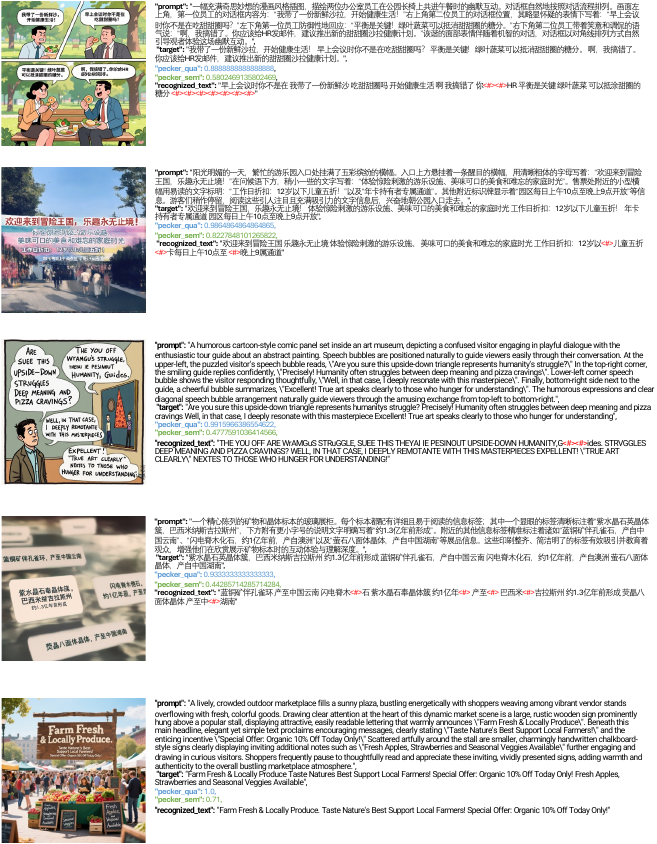}
\caption{This figure presents evaluation samples of TextPecker, showcasing its performance in detecting structural anomalies across diverse text rendering scenarios. $\omega = 1$ for structural quality score visualization.}
\label{supp:fig:textpecker_eval}
\end{figure*}

We provide additional visualizations of manually annotated structural anomalies from diverse generative models and synthetic structural anomalies generated by our rendering engine in Fig.~\ref{supp:fig:data_vis}, evaluation samples of TextPecker in Fig.~\ref{supp:fig:textpecker_eval}, and qualitative comparisons between Flux.1[dev] \cite{flux} and its RL-optimized variants in Fig.~\ref{fig:RL_EXP_Flux}.

\section{Additional Ablation Studies} 
\label{sup:sec:ablations}
\begin{table}[ht]
\scriptsize
\setlength{\tabcolsep}{0.9mm}  
\caption{Ablation study on the effectiveness of reward design:  PM for Pair-wise Matching, and SQ for Structural Quality reward, measured by TextPecker (InternVL3).}
\label{tab:rl_ablation_supp}
\renewcommand\arraystretch{1.1}
\centering
\begin{tabular}{l|c|ccc|cc}   
\toprule
\multirow{2}{*}{Generative Model} & \multirow{2}{*}{OCR Model} & \multicolumn{3}{c|}{Settings} & \multicolumn{2}{c}{GenTextEval-EN} \\
\cmidrule(lr){3-5} \cmidrule(lr){6-7}  
& & NED & PM & SQ & Qua. & Sem. \\
\midrule
\multirow{5}{*}{SD3.5-M \cite{SD3.5}} 
& - &   &  &  & 0.671 & 0.265 \\
& PP-OCRv5 \cite{PPOCRV5} &\checkmark   &  &  & 0.907 & 0.470 \\
& PP-OCRv5 \cite{PPOCRV5} & \checkmark  & \checkmark &  &  0.910 & 0.482 \\
& TextPecker & \checkmark  & \checkmark &  & 0.956 &     0.498   \\
& TextPecker & \checkmark   & \checkmark & \checkmark & \textbf{0.959} & \textbf{0.506} \\
\bottomrule
\end{tabular}
\vspace{-10pt}
\end{table}
We also provide additional ablation studies to deconstruct the reward function step-by-step and isolate the impact of each component, using StableDiffusion3.5-Medium \cite{SD3.5} as the baseline, as shown in Tab.~\ref{tab:rl_ablation_supp}. Combining a conventional, structure-unaware OCR model with Pairwise Matching (PM) yields a 1.2\% gain in semantic alignment, while structural quality sees a marginal 0.3\% improvement—indicating PM enhances semantic feedback but minimally boosts structural fidelity without structural perception. Replacing the OCR model with TextPecker delivers gains across both dimensions (Sem. +1.6\%, Qua. +4.6\%), demonstrating the value of our structure-aware assessor. Finally, incorporating the structural quality term as an auxiliary reward brings further improvements (Sem. +0.8\%, Qua. +0.3\%) and achieves the best overall performance, confirming the synergy of the full TextPecker reward design.
\section{Additional Generalization Results}
\begin{figure}[h]
  \centering
  \includegraphics[width=0.99\linewidth]{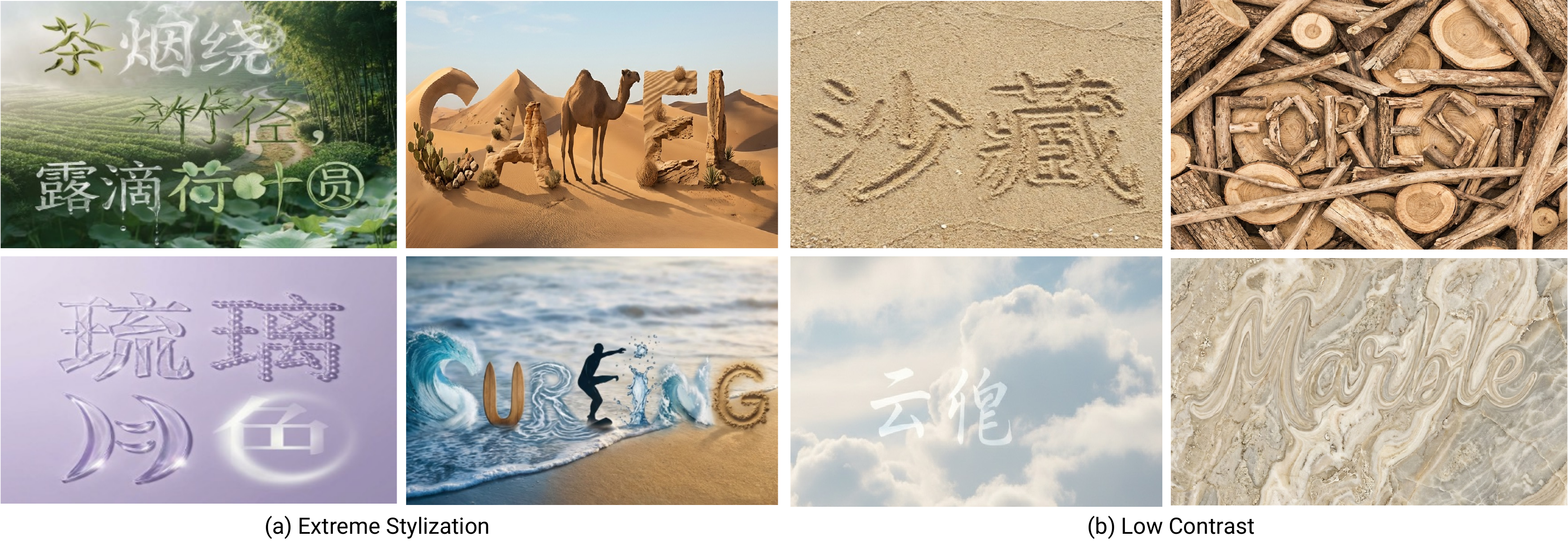}
   \caption{Hard Cases of TextPecker on Gemini-Rendered \cite{gemini2.5pro} Visual Text with Extreme Stylization and Low Contrast Layout.}
  \label{fig:hardcases}
\end{figure}

We conduct {\bf cross-model validation} on Gemini-2.5-flash-image \cite{gemini2.5pro} renderings to assess robustness under normal and extreme conditions (Tab.~\ref{tab:robustness_eval}), and the results are consistent across these settings, with failures mainly on extremely stylized fonts where artistic deformations distort canonical glyph structure and blur the boundary between style and true structural errors (see Fig.~\ref{fig:hardcases}, cases are all from Gemini). As for font variability, our dataset spans a large and diverse font pool across training and evaluation (Tab.~\ref{supp:tab:engine_params})

\vspace{-8pt}

\begin{table}[h]
\tiny
\setlength{\tabcolsep}{2.3mm}  
\caption{Robustness evaluation of TextPecker on Gemini-2.5-flash-image \cite{gemini2.5pro}: Performance under normal condition, extreme stylization, and low-contrast layouts.}
\label{tab:robustness_eval}
\renewcommand\arraystretch{1.1}
\centering
\begin{tabular}{l|cc|cc|cc}   
\toprule
\multirow{2}{*}{\textbf{Tab.~A} \& Methods} & \multicolumn{2}{c|}{Normal}  & \multicolumn{2}{c|}{Extreme Stylization} & \multicolumn{2}{c}{Low Contrast} \\ 
\cmidrule(lr){2-3} \cmidrule(lr){4-5} \cmidrule(lr){6-7}
& TSAP-F1 & CTR-R & TSAP-F1 & CTR-R & TSAP-F1 & CTR-R  \\ 
\midrule
InternVL-3-8B \cite{internvl3} & 0.000 & 0.666 & 0.087 & \textbf{0.588} & 0.364 & 0.742 \\ 
TextPecker-8B& \textbf{0.752} & \textbf{0.833} & \textbf{0.571} & \underline{0.577} & \textbf{0.800} & \textbf{0.839} \\ 
\bottomrule
\end{tabular}
\vspace{-13pt}
\end{table}

\begin{table}[h]
\scriptsize

\setlength{\tabcolsep}{0.9mm}

\caption{Additional Quantitative Comparisons of RL-Optimized Generative Models on \textbf{Chinese} Visual Text Benchxmarks (OneIG \cite{chang2025oneig}, LongText \cite{X-Omni_RL_and_LongText}, GenTextEval) with multi reward setting. \textbf{O:} OCR Reward \cite{flowgrpo}, \textbf{S:} TextPecker Semantic Reward, \textbf{Q:} TextPecker Structural Quality reward, \textbf{P:} Pickscore Reward \cite{pickscore}, \textbf{A:} Aethetic Reward \cite{laionaesthetics_score}. Results measurement and reward computation are both conducted by TextPecker (InternVL-3).}

\vspace{-6pt}

\label{tab:rl_chinese_benchmarks}

\renewcommand\arraystretch{0.85}

\centering


\begin{tabular}{l|c|c|cc|cc|cc}

\toprule

\multirow{2}{*}{method} & \multirow{2}{*}{rewards} & \multirow{2}{*}{weights}

& \multicolumn{2}{c|}{OneIG} & \multicolumn{2}{c|}{LongText} & \multicolumn{2}{c}{GenTextEval} \\

\cmidrule(lr){4-5}  \cmidrule(lr){6-7}  \cmidrule(lr){8-9}

& & & Qua. & Sem. & Qua. & Sem. & Qua. & Sem. \\

\midrule


\multirow{3}{*}{Qwen-Image \cite{QwenImage}}    & -- & -- & 0.888 & 0.747 & 0.900 & 0.815 & 0.922 & 0.805 \\

                               & OPA & 7:1:2 & 0.898 & 0.788 & 0.912 & 0.845 & 0.944 & 0.859 \\

& SQPA & 5:2:1:2 & \textbf{0.943} & \textbf{0.828} & \textbf{0.941} & \textbf{0.889} & \textbf{0.970} & \textbf{0.893} \\

\bottomrule

\end{tabular}

\vspace{-8pt}

\end{table}

\section{Additional Results on RL for VTR}
\label{supp:SecD:new_results}
To further validate the efficacy of TextPecker and attain more robust performance in Visual Text Rendering, we conduct additional experiments under \textbf{a strengthened RL baseline setting}, with key design choices elaborated as follows:

\noindent \textbf{Backbone enhancement.} We adopt recent GRPO-related techniques\cite{mixgrpo,grpoguard} to substantially enhance the efficiency and stability of the VTR optimization process, with implementation details supplemented in Sec.~\ref{sup:sec:impl_details}:

\noindent (i) Flow-GRPO-Fast \cite{mixgrpo} is employed to accelerate training convergence by injecting stochasticity only on partial optimization steps instead of all steps;

\noindent (ii) GRPO-Guard \cite{grpoguard} is employed to stabilize the training dynamics and mitigate implicit over-optimization issues in flow matching;

\noindent (iii) KL regularization enhancement (discussed in FlowGRPO’s \cite{flowgrpo} GitHub issues) is introduced to further alleviate over-optimization and reward hacking problems. 
The original formulation is:
\[
D_{\text{KL}}(\pi_\theta \parallel \pi_{\text{ref}}) = \frac{\left\| \bar{{x}}_{t+\Delta t,\theta} - \bar{{x}}_{t+\Delta t,\text{ref}} \right\|^2}{2\sigma_t^2 \Delta t}
\]

To stabilize training dynamics and mitigate over-optimization more effectively, we redefine the KL divergence to operate over \textbf{velocity-based} policy distributions:
\[
D_{\text{KL}}(\pi_\theta \parallel \pi_{\text{ref}}) = \left\| {v}_\theta({x}_t,t) - {v}_{\text{ref}}({x}_t,t) \right\|^2
\]

where all symbols follow the definitions in the FlowGRPO \cite{flowgrpo} paper, with the core adjustment being the switch from state ($\bar{{x}}$) to velocity (${v}$) as the regularization target.

\noindent \textbf{Multi-Reward Regularization.} In the main paper, we validated TextPecker’s efficacy via experiments exclusive to text-rendering rewards. However, this single-reward setup inevitably degrades the model’s aesthetic and image quality performance.
To yield more robust VTR optimization, we propose a multi-reward regularization strategy: we augment the original TextPecker reward with PickScore \cite{pickscore} and Aesthetic Score \cite{laionaesthetics_score}, implicitly regularizing the VTR model to yield more robust RL optimization results.

\begin{figure*}[t!]
  \centering
  \includegraphics[width=\linewidth]{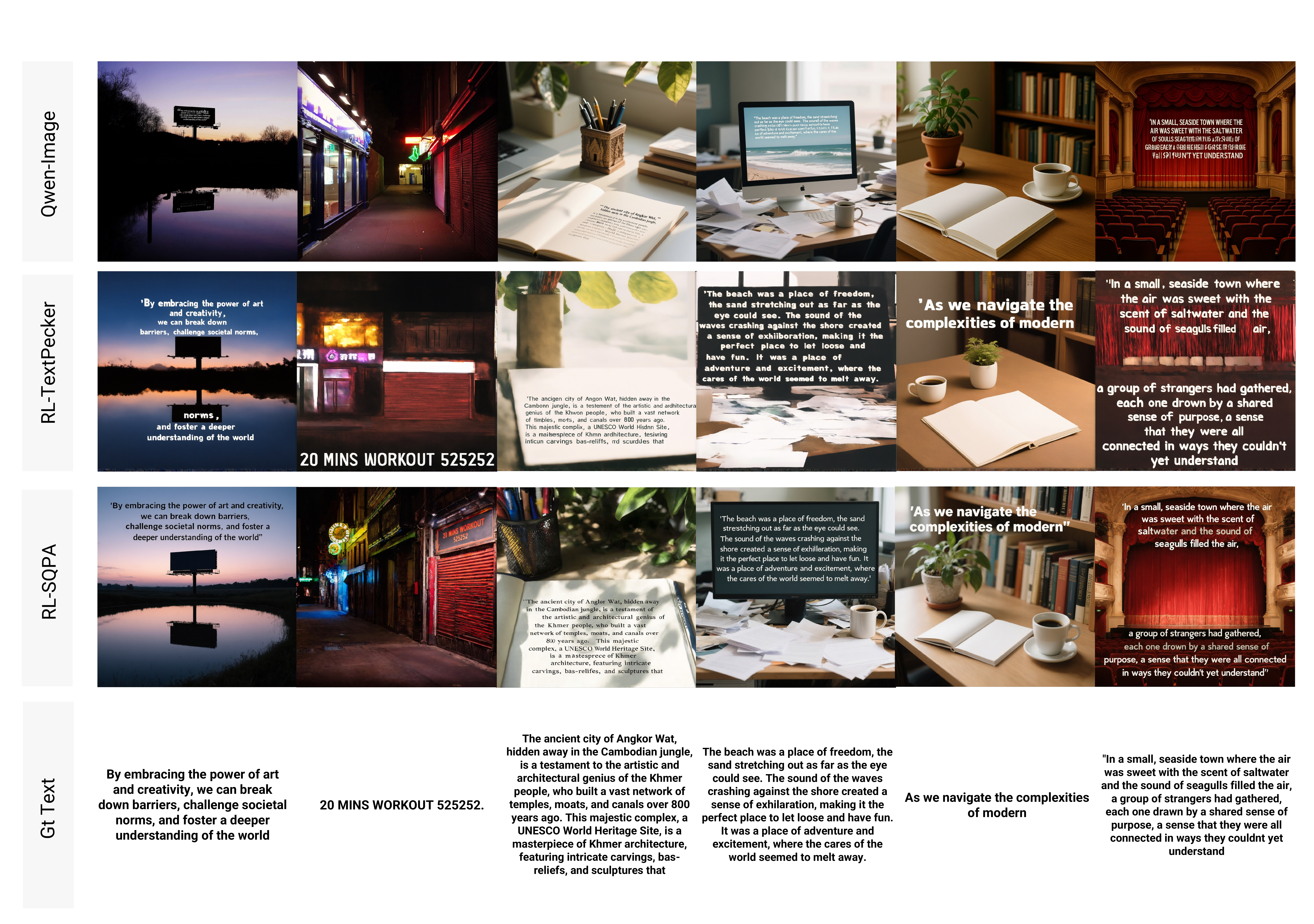}
  \vspace{-10pt}
  \caption{Qualitative comparisons of text rendering results (\textbf{English}) among different RL baseline settings.
  RL-TextPecker denotes the RL setting in the main paper,
  and RL-SQPA refers to our enhanced RL setting as described in Sec.~\ref{supp:SecD:new_results}.}
  \label{fig:RL_SQPA_qwen_en}
  \vspace{-10pt}
\end{figure*}

\begin{figure*}[t!]
  \centering
  \includegraphics[width=\linewidth]{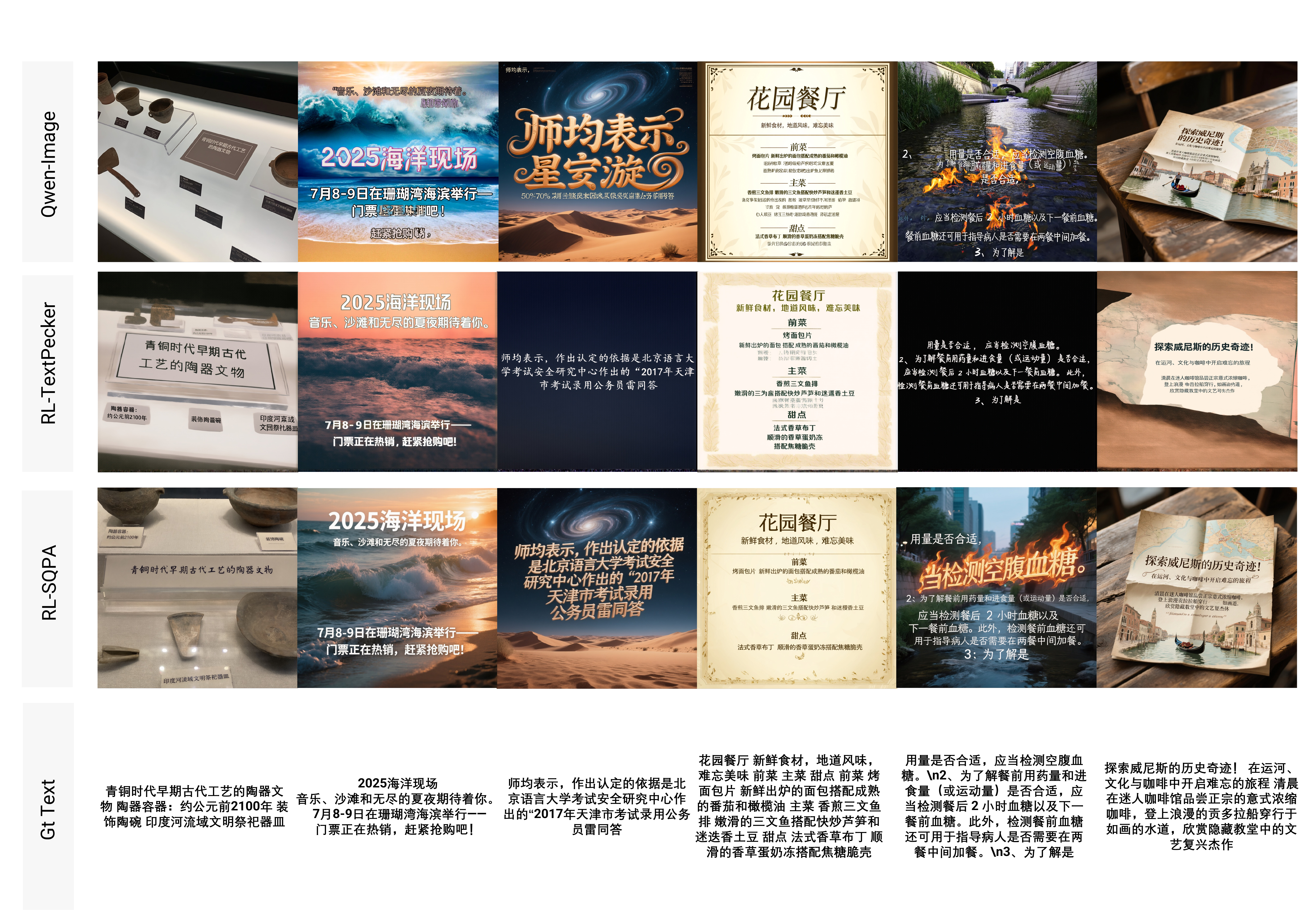}
  \vspace{-10pt}
  \caption{Qualitative comparisons of text rendering results (\textbf{Chinese}) among different RL baseline settings.
  RL-TextPecker denotes the RL setting in the main paper,
  and RL-SQPA refers to our enhanced RL setting as described in Sec.~\ref{supp:SecD:new_results}.}
  \label{fig:RL_SQPA_qwen_zh}
  \vspace{-10pt}
\end{figure*}

We present quantitative results of TextPecker under this enhanced RL baseline in Tab.~\ref{tab:rl_chinese_benchmarks} and Tab.~\ref{tab:rl_english_benchmarks}. Please note that all figures in the main paper and appendix are based on our original RL baseline,
except for the additional visual comparisons between the two TextPecker variants provided in Fig.~\ref{fig:RL_SQPA_qwen_en} and Fig.~\ref{fig:RL_SQPA_qwen_zh}.

\begin{table*}[ht]
\small
\setlength{\tabcolsep}{0.9mm}
\caption{Additional Quantitative Comparisons of RL-Optimized Generative Models on \textbf{English} VTR Benchmarks (OneIG \cite{chang2025oneig}, LongText \cite{X-Omni_RL_and_LongText}, CVTG \cite{TextCrafter_cvtg2k}, GenTextEval, TIIF \cite{TIIF-bench}, TextAtlas \cite{TextAtlas5M}, LeX \cite{Lex_art_10k}) with multi reward setting. \textbf{O:} OCR Reward \cite{flowgrpo}, \textbf{S:} TextPecker Semantic Reward, \textbf{Q:} TextPecker Structural Quality reward, \textbf{P:} Pickscore Reward \cite{pickscore}, \textbf{A:} Aethetic Reward \cite{laionaesthetics_score}. Results measurement and reward computation are both conducted by TextPecker (InternVL-3).}
\vspace{-6pt}
\label{tab:rl_english_benchmarks}
\renewcommand\arraystretch{1.0}
\centering
\begin{tabular}{l|c|c|cc|cc|cc|cc|cc|cc|cc}
\toprule
\multirow{2}{*}{method} & \multirow{2}{*}{rewards} & \multirow{2}{*}{weights}
& \multicolumn{2}{c|}{OneIG} & \multicolumn{2}{c|}{LongText} & \multicolumn{2}{c|}{CVTG} 
& \multicolumn{2}{c|}{GenTextEval} & \multicolumn{2}{c|}{TIIF} & \multicolumn{2}{c|}{TextAtlas}
& \multicolumn{2}{c}{LeX} \\
\cmidrule(lr){4-5} \cmidrule(lr){6-7} \cmidrule(lr){8-9} \cmidrule(lr){10-11} \cmidrule(lr){12-13} \cmidrule(lr){14-15} \cmidrule(lr){16-17}
& & & Qua. & Sem. & Qua. & Sem. & Qua. & Sem. & Qua. & Sem. & Qua. & Sem. & Qua. & Sem. & Qua. & Sem. \\
\midrule
\multirow{3}{*}{SD3.5-M \cite{SD3.5}}    & -- & -- & 0.840 & 0.507 & 0.836 & 0.407 & 0.843 & 0.466 & 0.666 & 0.262 & 0.758 & 0.347 & 0.646 & 0.269 & 0.810 & 0.454 \\
                      & OPA     & 7:1:2 & 0.908 & 0.588 & 0.913 & 0.508 & 0.895 & \textbf{0.621} & 0.896 & 0.461 & 0.886 & 0.483 & 0.916 & 0.436 & 0.894 & 0.563 \\
                      & SQPA    & 5:2:1:2 & \textbf{0.940} & \textbf{0.607} & \textbf{0.959} & \textbf{0.534} & \textbf{0.926} & 0.587 & \textbf{0.954} & \textbf{0.519} & \textbf{0.941} & \textbf{0.506} & \textbf{0.954} & \textbf{0.462} & \textbf{0.940} & \textbf{0.591} \\
\midrule
\multirow{3}{*}{Flux.1[dev] \cite{flux}}  & -- & -- & 0.870 & 0.578 & 0.925 & 0.584 & 0.889 & 0.510 & 0.664 & 0.332 & 0.933 & 0.540 & 0.683 & 0.307 & 0.946 & 0.667 \\
                      & OPA     & 7:1:2 & 0.977 & 0.739 & 0.977 & 0.763 & 0.974 & 0.780 & 0.982 & 0.739 & 0.986 & 0.719 & 0.983& 0.640 & 0.988 & 0.741 \\
                      & SQPA    & 5:2:1:2 & \textbf{0.990} & \textbf{0.775} & \textbf{0.992} & \textbf{0.780} & \textbf{0.993} & \textbf{0.824} & \textbf{0.991} & \textbf{0.762} & \textbf{0.991} & \textbf{0.735} & \textbf{0.993} & \textbf{0.649} & \textbf{0.991} & \textbf{0.807} \\
\midrule
\multirow{3}{*}{Qwen-Image \cite{QwenImage}}  & -- & -- & 0.954 & 0.812 & 0.961 & 0.831 & 0.960 & 0.817 & 0.958 & 0.723 & 0.933 & 0.682 & 0.953 & 0.665 & 0.927 & 0.760 \\
                      & OPA     & 7:1:2 & 0.963 & 0.840 & 0.967 & 0.858 & 0.962 & 0.848 & 0.974 & 0.808 & 0.964     & 0.764     & 0.970    & 0.728     & 0.958     & 0.850     \\
                      & SQPA    & 5:2:1:2 & \textbf{0.983} & \textbf{0.888} & \textbf{0.982} & \textbf{0.891} & \textbf{0.976} & \textbf{0.889} & \textbf{0.990 } & \textbf{0.876} & \textbf{0.975}     & \textbf{0.800}     & \textbf{0.982}    &  \textbf{0.746}    & \textbf{0.968}      & \textbf{0.883}     \\
\bottomrule
\end{tabular}
\vspace{-8pt}
\end{table*}

\section{Details of Dataset} 
\label{sup:sec:more_on_dataset}

\begin{table}[ht]
\scriptsize
\setlength{\tabcolsep}{2.4mm}
\caption{Statistics of English text-rich images generated by different models, labeled at box and image levels. Proportions are computed over all instances. }
\label{supp:tab:english_dataset_stats_with_overall_prop}
\renewcommand{\arraystretch}{1.1}
\centering
\begin{tabular}{c|c|c|c}
\toprule 
\textbf{Model}   & \textbf{Level} & \textbf{Samples} & \textbf{Proportion} \\
\midrule
\multirow{2}{*}{AnyText \cite{anytext}} & Box-level & 30647 & 7.38\% \\
                         & Image-level & 7405 & 1.78\% \\
\midrule
\multirow{2}{*}{Flux.1[dev] \cite{flux}} & Box-level & 91705 & 22.07\% \\
                      & Image-level & 19181 & 4.62\% \\
\midrule
\multirow{2}{*}{Qwen-Image \cite{QwenImage}} & Box-level & 20308 & 4.89\% \\
                             & Image-level & 2647 & 0.64\% \\
\midrule
\multirow{2}{*}{SD3 \cite{SD3.5}} & Box-level & 105725 & 25.45\% \\
                     & Image-level & 17766 & 4.28\% \\
\midrule
\multirow{2}{*}{SDv15 \cite{SD}} & Box-level & 61961 & 14.91\% \\
                       & Image-level & 6033 & 1.45\% \\
\midrule
\multirow{2}{*}{Seedream3.0 \cite{Seedream3.0}} & Box-level & 47921 & 11.53\% \\
                              & Image-level & 4144 & 1.00\% \\
\bottomrule
\end{tabular}
\vspace{-10pt}
\end{table}
\begin{table}[ht]
\scriptsize
\setlength{\tabcolsep}{2.4mm}
\caption{Statistics of Chinese text-rich images generated by different models, labeled at box and image levels. Proportions are computed over all instances. }
\label{supp:tab:chinese_dataset_stats_with_overall_prop}
\renewcommand{\arraystretch}{1.1}
\centering
\begin{tabular}{c|c|c|c}
\toprule 
\textbf{Model}   & \textbf{Level} & \textbf{Samples} & \textbf{Proportion} \\
\midrule
\multirow{2}{*}{CogView4 \cite{cogview}} & Box-level & 36000 & 11.14\% \\
                          & Image-level & 17005 & 5.26\% \\
\midrule
\multirow{2}{*}{Kolors \cite{Kolors}} & Box-level & 35549 & 10.99\% \\
                        & Image-level & 19312 & 5.98\% \\
\midrule
\multirow{2}{*}{Qwen-Image \cite{QwenImage}} & Box-level & 26597 & 8.23\% \\
                             & Image-level & 6225 & 1.93\% \\
\midrule
\multirow{2}{*}{Seedream3.0 \cite{Seedream3.0}} & Box-level & 146395 & 45.31\% \\
                              & Image-level & 36032 & 11.15\% \\
\bottomrule
\end{tabular}
\vspace{-10pt}
\end{table}

\subsection{Details on Text-rich Image Generation}
This section supplements the text-rich image generation dataset construction details in the main text. We present detailed statistics on the number of generated images categorized by language and various generative models employed.  English dataset statistics in Tab.~\ref{supp:tab:english_dataset_stats_with_overall_prop} and Chinese dataset statistics shown in Tab.~\ref{supp:tab:chinese_dataset_stats_with_overall_prop}.

\subsection{Details on Synthetic Data Augmentation}

As demonstrated in the main paper, models trained solely on manually annotated data generalize poorly to unseen structural anomalies. This limitation is particularly acute for Chinese characters, whose 2D structure and vast inventory (~8,000 common characters) create a combinatorial explosion of anomalies impossible to annotate exhaustively. To overcome this, we extend the SynthTIGER~\cite{synthtiger} renderer with two key enhancements: (1) image-level layout arrangements to simulate complex scenes, (2) Structural Anomaly Construction engine tailored to systematically generate diverse structural errors in Chinese.

Key parameters for our rendering engine, covering both canonical text generation and structural anomaly construction, are detailed in Table~\ref{supp:tab:engine_params}.
Our parameter choices are guided by the goal of training precise structural perception, not robust text extraction. Consequently, to preserve clear structural features, we intentionally disabled heavy post-processing (e.g., noise, blur) and certain style effects (e.g., extrusion), while limiting geometric transformations (e.g., skew, rotation) to moderate ranges. Notably, we have \textbf{a font pool of 976 types} to enhance font diversity. This ensures the rendered text maintains high structural clarity amidst realistic diversity, which is critical for our training objective.

\subsection{Structural Anomaly Perception Test Set}
\begin{table}[ht]
\scriptsize
\setlength{\tabcolsep}{2.4mm}
\caption{Statistics of our constructed text-rich image structural perception test dataset with structural-anomaly labels at box and image levels. Proportions are computed over all instances.}
\label{tab:test_set_stat}
\renewcommand{\arraystretch}{1.1}
\centering
\begin{tabular}{c|c|c|c}
\toprule 
Data Type      & Level   & Samples  & Proportion \\
\midrule

\multirow{2}{*}{Manual Annotations}          
& Box & 444    & 41.85\%     \\
& Image & 417        & 39.29\%    \\
\midrule 

\multirow{2}{*}{Synthetic Anomaly Text}            
& Box & 50    & 4.71\%     \\
& Image & 50        & 4.71\%    \\
\midrule 

\multirow{2}{*}{Synthetic Normal Text} 
& Box & 50    & 4.71\%     \\
& Image & 50        & 4.71\%    \\
\midrule

\multirow{1}{*}{\textbf{Total}} 
& \textbf{--}   &  1061    & 100\%   \\
\bottomrule

\end{tabular}
\end{table}


We provide detailed statistics of the structural perception test set in Tab.~\ref{tab:test_set_stat}.
To further validate the fairness and effectiveness of our results, we additionally conduct evaluations on a \textbf{real-only} test split (all synthetic samples excluded), with results presented in Tab.~\ref{tab:anno_only}. 
\begin{table}[h]
\tiny
\setlength{\tabcolsep}{0.8mm}
\centering
\renewcommand\arraystretch{0.85}
\caption{Performance of TextPecker on Real-only Test Splits}
\label{tab:anno_only}
\begin{tabular}{l|cccc|cccc}
\toprule
\multirow{3}{*}{\textbf{Methods}} & \multicolumn{4}{c|}{Chinese} & \multicolumn{4}{c}{English} \\
\cmidrule(lr){2-5} \cmidrule(lr){6-9}
& \multicolumn{2}{c|}{Image} & \multicolumn{2}{c|}{Box} & \multicolumn{2}{c|}{Image} & \multicolumn{2}{c}{Box} \\
\cmidrule(lr){2-3} \cmidrule(lr){4-5} \cmidrule(lr){6-7} \cmidrule(lr){8-9}
& TSAP-F1 & CTR-R & TSAP-F1 & CTR-R & TSAP-F1 & CTR-R & TSAP-F1 & CTR-R \\
\midrule
InternVL3-8B (Baseline) & \textcolor{red}{0.106} & \textbf{0.955} & \textcolor{red}{0.244} & 0.791 & \textcolor{red}{0.183} & 0.759 & \textcolor{red}{0.304} & 0.570 \\
TextPecker-8B (Anno) & \underline{0.866} & 0.849 & \underline{0.906} & \underline{0.815} & \textbf{0.874} & \textbf{0.938} & \underline{0.809} & \underline{0.918} \\
TextPecker-8B (Anno + Syn) & \textbf{0.901} & \underline{0.917} & \textbf{0.955} & \textbf{0.995} & \underline{0.850} & \underline{0.931} & \textbf{0.840} & \textbf{0.944} \\
\bottomrule
\end{tabular}
\end{table}

\subsection{Statistics of the RL Prompt Set for VTR}
\label{supp:RL_PROMPT}

\begin{figure}[hb!]
\centering
\includegraphics[width=\linewidth]{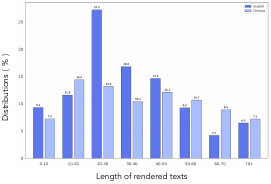}
\vspace{-15pt}
\caption{Statistics of the RL prompt set for RL-based VTR optimization (English: word-level; Chinese: character-level).}
\label{supp:fig:rl_data}
\end{figure}

We present the statistics of the curated prompt set used for RL-based VTR optimization in Fig.~\ref{supp:fig:rl_data}. The prompt set is designed to encompass diverse text lengths and content for effective reinforcement learning. 

For English text rendering, we curate prompts from TextAtlas5M \cite{TextAtlas5M}, ensuring a rich and varied dataset. For Chinese text rendering, we adopt a similar paradigm as described in the main paper, starting with a comprehensive text corpus sampled from WanJuan1.0 \cite{wanjuan}, which covers a wide range of modern Chinese common characters. Additionally, we use Qwen3-235B-A22B \cite{qwen3-235b-a22b-ins} to generate diverse style descriptions of fonts. These style descriptions are integrated with the corpus to create the final prompt set. The statistics are visualized in Fig.~\ref{supp:fig:rl_data}.

\subsection{Statistics of the GenTextEval Dataset}

\begin{figure*}[ht]
  \centering
  \includegraphics[width=0.75\linewidth]{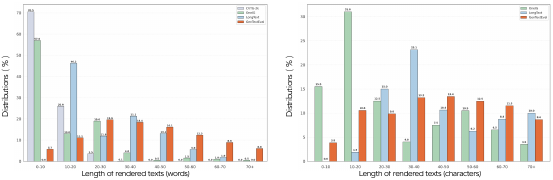}
  \vspace{-5pt}
  \caption{Comparison among CVTG-2K\cite{TextCrafter_cvtg2k}, OneIG-Bench\cite{chang2025oneig}, LongText-Bench\cite{X-Omni_RL_and_LongText}, and Our Proposed GenTextEval-Bench with Respect to the Length of Rendered Texts in English (Left) and Chinese (Right).}
  \label{supp:fig:gentexteval}
\end{figure*}

To facilitate re-evaluation with TextPecker and build upon the strengths of existing benchmarks, we construct a dataset named GenTextEval, which integrates English and Chinese prompts from multiple sources \cite{chang2025oneig,X-Omni_RL_and_LongText,TextCrafter_cvtg2k,TextAtlas5M}. In light of the limited availability of Chinese-rendering benchmarks \cite{chang2025oneig,X-Omni_RL_and_LongText,QwenImage} (with ChineseWord remaining unavailable for open-source at the time of our experiments), this dataset is further enriched with Chinese prompts curated as described in Sec.~\ref{supp:RL_PROMPT}. The final GenTextEval dataset comprises 314 prompts for English rendering and 417 prompts for Chinese rendering. Following the traditional paradigm, each prompt generates four distinct image outputs to ensure fairer assessment.
We offer a statistical overview of the GenTextEval dataset in Fig.~\ref{supp:fig:gentexteval}.

\section{Prompt Template for TextPecker}

\begin{figure*}[t!]
\centering
\includegraphics[width=0.8\linewidth]{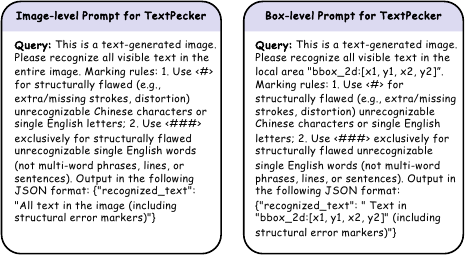}
\caption{The prompting template used for our TextPecker.}
\label{supp:fig:prompt_template}
\end{figure*}

We present the prompting templates used for TextPecker’s training and testing in Fig.~\ref{supp:fig:prompt_template}. To ensure consistency and comparability across evaluations, we adopt the identical template for all other MLLM baselines.

\section{Additional Implementation Details}
\label{sup:sec:impl_details}

This section provides further implementation details, supplementing the overview in the main paper.
We employ the Flow-GRPO \cite{flowgrpo} framework for all Reinforcement Learning (RL) based VTR optimization experiments. Notably, Flow-GRPO is an actively evolving repository, and the implementations reported here reflect the stable version available at the time of our experiments. As noted in the main paper, the overall methodology adheres to Flow-GRPO’s core design, while the specific hyperparameters are carefully tuned for each base model (building upon the framework’s default configurations) to ensure stable and effective training.  The resolution for all generated images is set to $512 \times 512$ pixels.
The model-specific configurations are detailed below.

\vspace{2mm} 
\noindent\textbf{SD3.5-M \cite{SD3.5}:}
We use 30 sampling steps for training and 40 for evaluation. The noise level is set to 0.8, and the guidance scale is 1.0 (following the Flow-GRPO-Fast framework, CPS sampling and No-CFG are adopted to improve training efficiency). The KL ratio $\beta$ is 0.04. For LoRA, we adopt a rank $r$ of 32 and an alpha $\alpha$ of 64. Training is conducted using the Flow-GRPO-Fast framework.

\vspace{2mm}
\noindent\textbf{Flux.1[dev] \cite{flux}:}
We set 14 sampling steps for training and 28 for evaluation. The noise level is 0.9, the guidance scale is 3.5, and the KL ratio $\beta$ remains 0.04. For LoRA, we use a rank $r$ of 64 and an alpha $\alpha$ of 128. No Flow-GRPO variant is employed for this model.

\vspace{2mm}
\noindent\textbf{Qwen-Image \cite{QwenImage}:}
We employ 10 sampling steps for training and 50 for evaluation. The noise level is set to 1.2, with a guidance scale of 4.0 and a KL ratio $\beta$ of 0.004. For LoRA, we adopt a rank $r$ of 64 and an alpha $\alpha$ of 128. Training is conducted using the Flow-GRPO-Fast framework.

\subsection{Computational cost and latency.}
The evaluator is used only during RL training and run as a separate asynchronous service, hence it adds negligible overhead and does not affect inference latency; on SD3.5-M\cite{SD3.5}, 100 RL steps take 5.52 h (TextPecker) vs.\ 5.40 h (PPOCRv5\cite{PPOCRV5}).




\section{Additional Implementation Details on Sec.~\ref{supp:SecD:new_results}}
\label{sup:sec:impl_details_2}

As mentioned in Sec.~\ref{supp:SecD:new_results}, we conducted additional experiments utilizing an enhanced RL baseline. This baseline incorporates several advanced techniques including Flow-GRPO-Fast \cite{mixgrpo}, GRPO-Guard \cite{grpoguard}, Velocity KL loss, and multi-reward regularization. This section provides the specific hyperparameter details for experiments in Tab.~\ref{tab:rl_chinese_benchmarks} and Tab.~\ref{tab:rl_english_benchmarks}. LoRA configurations remain identical to those described in Sec.~\ref{supp:SecD:new_results}.

\vspace{2mm}
\noindent\textbf{SD3.5-M \cite{SD3.5}:}
We use 40 sampling steps for both training and evaluation. An SDE window of size 12 is applied during the first half of the sampling process. The key hyperparameters are set as follows: a noise level of 0.9, a guidance scale of 4.5, and a learning rate of $10^{-4}$. The ratio $\beta$ for the Velocity KL loss is set to $10^{-4}$, and the clipping range is set to $2 \times 10^{-6}$.

\vspace{2mm}
\noindent\textbf{Flux.1[dev] \cite{flux}:}
We set the number of sampling steps to 28 for both training and evaluation. An SDE window of size 9 is used in the first half of the sampling steps. The noise level is 0.9, the guidance scale is 3.5, and the learning rate is $10^{-4}$. The Velocity KL loss ratio $\beta$ is configured to $10^{-4}$, and the clipping range is set to $2 \times 10^{-6}$.

\vspace{2mm}
\noindent\textbf{Qwen-Image \cite{QwenImage}:}
We employ 20 sampling steps for training and 50 for evaluation. During training, an SDE window of size 5 is applied to the initial half of the sampling steps. The noise level is set to 1.2, the guidance scale is 4, and the learning rate is $10^{-4}$. The Velocity KL ratio $\beta$ is $10^{-3}$, and the clipping range is set to $2 \times 10^{-5}$.
\section{Limitations}
\label{sup:sec:Limitations}
TextPecker paves a novel path for addressing the core bottleneck in VTR evaluation and RL-based optimization, leveraging a structural-anomaly-aware RL reward that delivers complementary signals for semantic alignment and structural quality. While providing a foundational step towards structurally faithful VTR, our work still has several limitations that point to meaningful directions to be explored.

\textit{First}, our structural anomaly synthesis is contingent upon the availability of stroke-level font data. This dependency currently restricts its application to standard fonts, precluding the generation of anomalies in artistic or proprietary typefaces lacking such data.

\textit{Second}, our work is currently confined to Chinese and English text rendering, with efficient multilingual extension as a key area for future exploration. 

\textit{Third}, our TextPecker evaluator is equipped with box-level perception ability, which theoretically enables it to support downstream VTR-related tasks such as text translation and local text editing, which are often challenging for general editing methods \cite{flux,Nano-Banana,FreeFine,shi2024seededit,anytext}. Validating the effectiveness of evaluation and RL optimization on these downstream tasks is left for future work.

\textit{Fourth}, challenges arise in handling artistic text generation (see Fig.~\ref{fig:hardcases} above), which is an increasingly demanded scenario. Artistic text often involves deliberate modifications to standard structures, such as connected strokes, added symbols, or pictorial variations, making it inherently difficult to define a single standard or ground truth. 
Furthermore, artistic designs are continuously evolving, presenting a moving target that conflicts with the structural consistency objectives of our current framework. Addressing the evaluation and optimization of artistic text generation remains a challenging yet impactful research direction, necessitating the integration of creative expression with principles of structure-aware textual modeling.

\begin{table*}[t]
\scriptsize
\setlength{\tabcolsep}{1.6mm}
\caption{Key parameters for canonical text rendering and structural anomaly construction.}
\label{supp:tab:engine_params}
\renewcommand{\arraystretch}{1.3}
\centering
\begin{tabular}{l|c|c|c}
\toprule 
\textbf{Parameter Category} & \textbf{Canonical Chinese Text} & \textbf{Canonical English Text} & \textbf{Structural Anomaly Construction} \\
\midrule
\multicolumn{4}{l}{\textbf{Basic Text Configuration}} \\
Vertical text probability & 10\% & 10\% & 10\% \\
Number of elements per sample & 3–10 & 3–10 & 3–10 \\
Text length range & 1–25 characters & 3–25 characters & 1–25 characters \\
\midrule
\multicolumn{4}{l}{\textbf{Font Settings}} \\
Number of font types & 976 & 976 & 976 \\
Font size range & 50–100 pt & 50–100 pt & 50–100 pt \\
\midrule
\multicolumn{4}{l}{\textbf{Layout Parameters}} \\
Horizontal spacing between elements & 50–200 px & 50–200 px & 50–200 px \\
Vertical line spacing & 10–20 px & 10–20 px & 10–20 px \\
Length ratio range & 0.8–1.0 & 0.8–1.0 & 0.8–1.0 \\
Random offset probability & 20\% & 20\% & 20\% \\
Random offset range & 10–30 px & 10–30 px & 10–30 px \\
Image margin & 15 px & 15 px & 15 px \\
Flow layout probability & 80\% & 80\% & 80\% \\
Curve layout probability & 20\% & 20\% & 20\% \\
\midrule
\multicolumn{4}{l}{\textbf{Style Effects}} \\
Style application probability & 25\% & 25\% & 25\% \\
- Text border (probability) & 100\% & 100\% & 100\% \\
  \quad Size ratio & 5–15\%  & 5–15\%  & 5–15\% \\
  \quad Alpha & 1.0 & 1.0 & 1.0 \\
- Text shadow (probability) & 0\% & 0\% & 0\% \\
- Text extrusion (probability) & 0\% & 0\% & 0\% \\
\midrule
\multicolumn{4}{l}{\textbf{Geometric Transformation}} \\
Transformation application probability & 50\% & 50\% & 50\% \\
- Perspective x (weight) & 1 & 1 & 1 \\
  \quad Percents & 0.8 & 0.8 & 0.8 \\
- Perspective y (weight) & 1 & 1 & 1 \\
  \quad Percents & 0.8–1 & 0.8–1 & 0.8–1 \\
- Trapezoidate x (weight) & 1 & 1 & 1 \\
  \quad Percent & 0.8–1 & 0.8–1 & 0.8–1 \\
- Trapezoidate y (weight) & 1 & 1 & 1 \\
  \quad Percent & 0.8–1 & 0.8–1 & 0.8–1 \\
- Skew x (weight) & 2 & 2 & 2 \\
  \quad Angle & 0–30° & 0–30° & 0–30° \\
- Skew y (weight) & 2 & 2 & 2 \\
  \quad Angle & 0–10° & 0–10° & 0–10° \\
- Rotate (weight) & 3 & 3 & 3 \\
  \quad Angle & 0–10° & 0–10° & 0–10° \\
\midrule
\multicolumn{4}{l}{\textbf{Structural Anomaly Generation}} \\
Anomaly generation probability & 0\% & 0\% & 50\% \\
- Deletion (probability) & – & – & 40\% \\
- Insertion (probability) & – & – & 40\% \\
- Swapping (probability) & – & – & 40\% \\
\bottomrule
\end{tabular}
\vspace{-10pt}
\end{table*}

\end{document}